%% file: acl_latex.tex
\documentclass[11pt]{article}

\usepackage[preprint]{acl}

\usepackage{times}
\usepackage{latexsym}

\usepackage[T1]{fontenc}
\usepackage[utf8]{inputenc}

\usepackage{microtype}

\usepackage{inconsolata}

\usepackage{graphicx}
\usepackage{booktabs}
\usepackage{multirow}
\usepackage{pifont}
\usepackage[table]{xcolor}
\newcommand{\cmark}{\ding{51}}
\newcommand{\pmark}{$\triangle$}
\newcommand{\xmark}{\ding{55}}
\usepackage[most]{tcolorbox}
\tcbuselibrary{listings,breakable}

\title{\textsc{Agora}: An Archive-Grounded Benchmark for Agentic\\
Workplace Document Reasoning}

\author{ \textbf{Honglin Guo\textsuperscript{1,}\thanks{Equal contribution.}}, \textbf{Qi Zhang\textsuperscript{1,}\footnotemark[1]}, \textbf{Yu Zhang\textsuperscript{2}}, \textbf{Weijie Li\textsuperscript{1}}, \textbf{Rui Zheng\textsuperscript{3}}, \\
\textbf{Zhikai Lei\textsuperscript{3}}, \textbf{Qiyuan Peng\textsuperscript{1}}, \textbf{Zhiheng Xi\textsuperscript{1}}, \textbf{Tao Gui\textsuperscript{1}}, \textbf{Qi Zhang\textsuperscript{1}} \\
\textsuperscript{1}Fudan University, \textsuperscript{2}Zhejiang University, \textsuperscript{3}Shanghai Qiji Zhifeng Co., Ltd. \\
hlguo26@m.fudan.edu.cn, tgui@fudan.edu.cn \\
\url{https://github.com/KYLN24/Agora} \\
\url{https://huggingface.co/datasets/KYLN24/Agora}
} 

\begin{document}
  \maketitle
  \begin{abstract}
    Large language models are increasingly deployed as agents that reason over documents rather than answer from parametric knowledge. We study \emph{archive-grounded reasoning}: locating sparse evidence across a large, messy collection of workplace files, reconciling inconsistent terminology, units, and time conventions, and computing an answer. Existing benchmarks address only parts of this setting and none jointly stresses archive-groundedness, agentic exploration, and cross-domain coverage. We introduce \textbf{\textsc{Agora}}\footnote{\textsc{Agora} is short for \textbf{A}rchive-\textbf{G}rounded \textbf{O}ffice \textbf{R}easoning \textbf{A}ssessment}, a benchmark pairing 362 questions with eight domain collections of 9{,}664 authentic documents and 372M tokens, far exceeding any model's context window, so agents must explore deliberately rather than scan exhaustively. \textsc{Agora} is built by an agentic pipeline combining cross-document task synthesis, leakage-preventing obfuscation, and difficulty filtering. Evaluating eight models, we find the task far from solved: even the strongest reaches only 59.4\% accuracy, with notable variation across domains.
  \end{abstract}

  \section{Introduction}

  \begin{figure}[t]
    \centering
    \includegraphics[width=\linewidth]{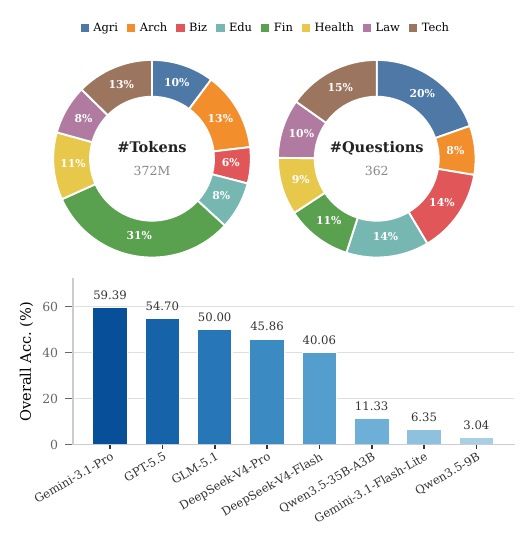}
    \caption{\textbf{Overview of \textsc{Agora}.} The benchmark comprises 372M tokens across 9{,}664 documents and 362 questions, spanning eight professional domains. Even the strongest model reaches only 59.4\% overall accuracy, leaving substantial headroom.}
    \label{fig:head}
  \end{figure}

  \input{tables/benchmark_comparison.tex}

  Large language models are increasingly being used as agents rather than standalone chatbots~\citep{DBLP:conf/nips/YangJWLYNP24,DBLP:journals/corr/abs-2503-21460,deepseek2026deepseekv4techreport}. In enterprise settings, their value is not limited to producing fluent responses, but lies in helping users complete real knowledge-work tasks over internal archives~\citep{DBLP:journals/corr/abs-2503-13964,DBLP:journals/corr/abs-2601-05163}. For example, financial analysts need to verify figures across reports and spreadsheets, legal researchers need to trace relevant clauses and precedents, and policy teams need to synthesize evidence from scattered documents. These tasks are difficult because internal archives are often large, inconsistently organized, and full of conflicting terminology, units, dates, and assumptions~\citep{DBLP:journals/corr/abs-2311-11944,DBLP:conf/acl/ZhaoLLZ22}. A capable agent must locate sparse evidence, reconcile inconsistencies, perform necessary calculations, and produce an answer that is accurate, verifiable, and directly useful for decision-making~\citep{DBLP:journals/corr/abs-2503-09516,DBLP:conf/emnlp/ZhengFHCYLL25}. Evaluating language agents therefore requires assessing not their fluency in chat but their capacity for archive-grounded document reasoning, the capability central to whether they can become reliable workplace assistants.

  As Table~\ref{tab:benchmark_comparison} shows, existing benchmarks cover only part of the requirements outlined above. Multi-hop QA benchmarks~\citep{DBLP:conf/emnlp/Yang0ZBCSM18,DBLP:journals/tacl/TrivediBKS22,DBLP:conf/naacl/KrishnaKMSSUF25} typically rely on homogeneous corpora such as Wikipedia, and therefore do not capture the file-format diversity and irregular organization of workplace archives. Document QA and table QA benchmarks~\citep{DBLP:conf/acl/ZhuLHWZLFC20,DBLP:journals/corr/abs-2311-11944} are usually framed as reading-comprehension tasks, so they do not require agents to navigate file systems or perform multi-step computation over heterogeneous documents. Agent and web-browsing benchmarks~\citep{DBLP:conf/iclr/MialonF0LS24,DBLP:conf/iclr/ZhouX0ZLSCOBF0N24,DBLP:journals/corr/abs-2504-12516} mostly operate on the open web or in simulated environments, rather than on a bounded internal archive. The closest prior work is OfficeQA Pro~\citep{DBLP:journals/corr/abs-2603-08655}, which combines retrieval with computation over a large enterprise-style corpus; however, it is built from a single external source, limiting its domain and file-format coverage. This gap motivates the need for a benchmark that evaluates whether LLM agents can reason over realistic internal archives rather than over flat, homogeneous, or web-scale corpora.

  \begin{figure*}[t]
    \centering
    \includegraphics[width=\textwidth]{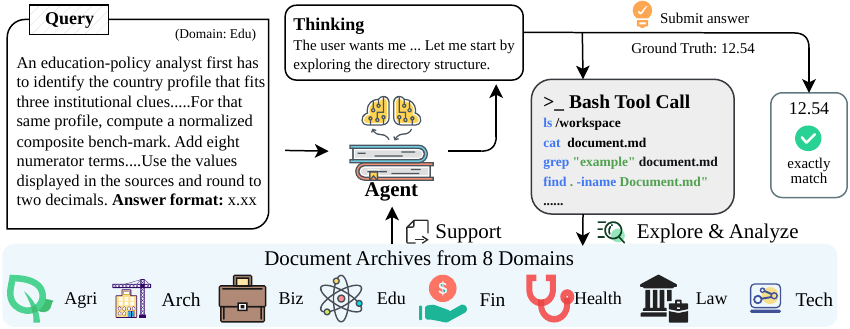}
    \caption{Overview of the \textsc{Agora} task setting. Given a natural-language query paired with one of eight professional domain collections, an agent explores files and runs computation through bash tool calls. After turns of exploration and analysis, it submits a single numeric answer that is deterministically verified against the gold answer.}
    \label{fig:overview}
  \end{figure*}

  To bridge this gap, we introduce \textbf{\textsc{Agora}}, a benchmark designed to test whether LLM agents can perform archive-grounded reasoning in realistic workplace settings. Real archive-grounded reasoning is not a simple reading-comprehension task over a flat corpus; it is a search-and-verify process under scarcity. \textsc{Agora} contains 362 natural-language queries distributed across eight domain-specific collections, covering 9{,}664 real-world documents and 372M tokens. For each query, an agent is given access only to its paired collection and must navigate the file structure, locate sparse evidence, and execute the computations needed to derive the answer. Because each collection is far larger than the context window of current models, the agent cannot rely on exhaustive scanning or shallow keyword matching; it must plan its exploration and reason over evidence selected from the archive. Each query has a unique, verifiable numeric answer, enabling deterministic and reproducible evaluation without human or model-based judgment. \textsc{Agora} is constructed through an agentic pipeline that collects documents across multiple domains using deep search, synthesizes cross-document multi-hop questions, and enforces quality through leakage-preventing obfuscation, difficulty filtering, and human verification. As Figure~\ref{fig:head} shows, the resulting benchmark remains highly challenging: across eight evaluated models, even the strongest reaches only 59.4\% accuracy.

  We evaluate a mix of eight proprietary and open-weight models on \textsc{Agora}, running all of them in \texttt{mini-swe-agent}, a minimal harness that exposes only a bash tool. This keeps the agent interface fixed and simple, helping focus the comparison on model-level reasoning, evidence selection, and tool use rather than on complex agent scaffolding. Archive-grounded document reasoning remains far from solved: even the strongest model obtains only 59.4\% accuracy, just below the 60\% threshold. More importantly, performance varies substantially across domains. The results show that difficulty is often not a property of the benchmark alone, but emerges from the interaction between a specific model and a specific domain. Indeed, the per-domain leaderboards often differ from the aggregate ranking, revealing weaknesses that would be hidden by a single-source or single-domain benchmark.

  Our contributions are threefold. First, we introduce \textbf{\textsc{Agora}}, a cross-domain benchmark for archive-grounded agentic document reasoning, consisting of 362 questions with verifiable numeric answers over eight workplace document collections, including 9{,}664 real-world documents and 372M tokens. Second, we develop an agentic construction pipeline that combines cross-document task synthesis, leakage-preventing obfuscation, difficulty filtering, and human verification. Third, we evaluate a mix of eight frontier proprietary and open-weight models on \textsc{Agora}, showing that archive-grounded document reasoning remains far from solved: even the strongest model reaches only 59.4\% accuracy and exhibits systematic per-domain weaknesses.

  \section{Related Work}

  The shift from standalone chatbots to autonomous agents has been driven by methodological advances~\citep{DBLP:conf/iclr/YaoZYDSN023,DBLP:conf/nips/SchickDDRLHZCS23}, which established the paradigm of interleaving reasoning with tool invocation~\citep{DBLP:conf/nips/YangJWLYNP24,DBLP:conf/iclr/0001LSXTZPSLSTL25}. Building on these foundations, agents are now widely deployed in real productivity settings, spanning code repair, end-to-end office workflows, and deep research over document collections~\citep{DBLP:journals/corr/abs-2603-15726,DBLP:journals/corr/abs-2505-15117}. As such agents proliferate, rigorous and scenario-faithful benchmarks become essential for measuring their real-world capability. A first thread of benchmarks evaluates agents in increasingly realistic environments. GAIA~\citep{DBLP:conf/iclr/MialonF0LS24} and BrowseComp~\citep{DBLP:journals/corr/abs-2504-12516} probe open-web reasoning with browsing and search tools, while WebArena~\citep{DBLP:conf/iclr/ZhouX0ZLSCOBF0N24}, DeepSynth~\citep{DBLP:journals/corr/abs-2602-21143} and OSWorld~\citep{DBLP:conf/nips/XieZCLZCHCSLLXZ24} target simulated browser and desktop operating-system environments. Closer to office workflows, SpreadsheetBench~\citep{DBLP:conf/nips/MaZZYZZLW024} evaluates cell-level manipulation and formula reasoning, MEBench~\citep{DBLP:conf/emnlp/LinLZZLWT25} measures multi-step tool use over office artifacts, and OfficeBench~\citep{DBLP:journals/corr/abs-2407-19056} extends the setting to task completion across diverse office software. Across this thread, evaluation is grounded in interactive environments or concrete office artifacts, with documents typically serving as targets of manipulation rather than as a body of evidence to be reasoned over.

  Closer to our setting, multi-document question answering over workspace document archive has attracted substantial attention. HotpotQA~\citep{DBLP:conf/emnlp/Yang0ZBCSM18}, 2WikiMultihopQA~\citep{DBLP:conf/coling/HoNSA20}, MuSiQue~\citep{DBLP:journals/tacl/TrivediBKS22}, and FRAMES~\citep{DBLP:conf/naacl/KrishnaKMSSUF25} require composing facts across multiple Wikipedia passages, often with explicit supporting-fact supervision; MultiHop-RAG~\citep{DBLP:journals/corr/abs-2401-15391} extends the setting to news corpora, and M3SCIQA~\citep{DBLP:conf/emnlp/LiS0L0C24} pushes multi-hop reasoning into the scientific literature. Closer to professional settings, TAT-QA~\citep{DBLP:conf/acl/ZhuLHWZLFC20} and FinanceBench~\citep{DBLP:journals/corr/abs-2311-11944} evaluate hybrid text-and-table reasoning over financial materials, GDPVal~\citep{DBLP:journals/corr/abs-2510-04374} measures economically valuable deliverables across occupations, and OfficeQA Pro~\citep{DBLP:journals/corr/abs-2603-08655} couples retrieval with computation over a specific source of corpus. None of these benchmarks, however, jointly stresses active retrieval over a large internal collection and cross-file reconciliation of units, time conventions, and terminology before computation—the conditions that define real productivity workflows and that motivate \textsc{Agora}.

  \section{The \textsc{Agora} Benchmark}

  Building a benchmark that jointly demands archive-groundedness, agentic exploration, and cross-domain coverage poses three challenges: specifying tasks that genuinely require reasoning over a fixed collection rather than parametric knowledge or single-file lookup; assembling authentic, messy documents at a scale that forces deliberate exploration; and synthesizing verifiable multi-hop questions while suppressing evidence leakage. We address these in turn: formalizing the task and benchmark and the pipeline that constructs it.

  \subsection{Dataset Desiderata}
  \label{sec:desiderata} We distill the design of \textsc{Agora} into four desiderata, derived from how document-grounded reasoning arises in real workplace settings and from our aim of a rigorously measurable benchmark.

  \paragraph{Archive-groundedness.}
  Each task must be answerable using only a fixed source collection $\mathcal{C}$, without open-web access, so that a score depends on the collection rather than a model's prior knowledge. A frozen collection also makes the benchmark reproducible: the available evidence does not drift over time.

  \paragraph{Cross-domain coverage.}
  The benchmark must span a broad range of professional domains. Real corpora differ sharply across domains in file formats, table structures, terminology, and reporting conventions, and a single-source benchmark risks rewarding agents that overfit to one source's idiosyncrasies. Drawing collections from many domains instead measures whether document-reasoning ability generalizes.

  \paragraph{Agentic exploration and evidence integration.}
  A task must test an agent end to end: planning its exploration, gathering a long-range evidence chain across files, and reconciling that evidence into an answer. Evidence is sparsely distributed among a large volume of unrelated material, so an agent must navigate deliberately rather than scan exhaustively, which is a particular challenge under a limited context window. Retrieved evidence moreover often fails to line up, differing in wording, definitions, or unit and time conventions, and these inconsistencies must be resolved before the final computation.

  \paragraph{Verifiable evaluation.}
  Every task must admit automatic and deterministic verification. To this end, each query has a single numeric answer and a specified output format, so responses can be checked against ground truth by normalizing superficial formatting differences without human or model-based judging, and free of its noise and cost.

  \medskip
  \noindent
  Together, these desiderata scope \textsc{Agora} deliberately: by pairing realistic, messy document environments with single numeric answers and automatic verification, it targets agentic exploration and reasoning that can be measured precisely, rather than the open-ended deliverable quality assessed by other benchmarks.

  \begin{figure*}[t]
    \centering
    \includegraphics[width=\textwidth]{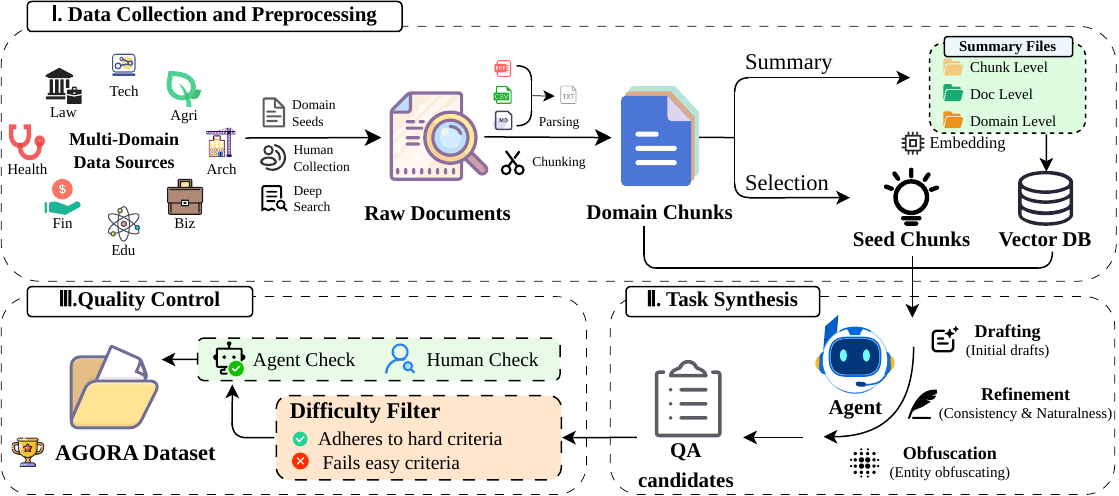}
    \caption{Overview of the \textsc{Agora} construction pipeline. Phase~1 (Data Collection and Preprocessing) gathers and parses multi-domain documents, segmenting them into chunks indexed in a vector database. Phase~2 (Task Synthesis) drafts tasks from these chunks and progressively enhances them through refinement and obfuscation. Phase~3 (Quality Control) applies difficulty filtering and multi-check verification to yield the final QA set.}
    \label{fig:construction_pipeline}
  \end{figure*}

  \subsection{Task and Benchmark Composition}
  \label{subsec:composition}

  \input{tables/benchmark_composition.tex}

  \paragraph{Task formulation.}

  Each \textsc{Agora} task pairs a natural-language query with one domain collection that the agent may use exclusively to answer it. A query is cross-document and multi-hop: its evidence is sparsely scattered across several files amid much unrelated material, so the agent must locate bridging facts, reconcile inconsistent terminology, units, and time conventions, and compute an answer. As illustrated in Figure~\ref{fig:overview}, answering such a query requires multiple rounds of tool calls to explore and analyze the corresponding document collection before submitting an answer in the specified format. Because each collection far exceeds any model's context window, the agent cannot scan it exhaustively but must explore deliberately. Every query admits a single verifiable numeric answer under a specified output format; the agent works through a bash tool and submits its answer in an \texttt{<answer>}\,\dots\,\texttt{</answer>} tag. Appendix~\ref{secapp:task_example} gives more task examples.

  \paragraph{Benchmark composition.}
  \textsc{Agora} consists of eight source collections, one per domain, each paired with a set of natural language queries answered over that collection. A collection is a flat directory of plain-text Markdown files converted from authentic workplace documents such as official reports, statistical yearbooks, and tabular records. Files are named after their originating data source and document title, and a single collection aggregates documents from several distinct data sources. The documents are authentic and may appear in languages other than English, while all queries are posed in English. In total, the benchmark comprises 362 questions. Table~\ref{tab:composition} reports the per-domain composition of \textsc{Agora}.

  \subsection{Benchmark Construction}
  \label{sec:construction}

  Figure~\ref{fig:construction_pipeline} illustrates the three-stage construction of \textsc{Agora}: document collection and preprocessing, task synthesis, and quality control. A \texttt{Codex}-based agent\footnote{\url{https://developers.openai.com/codex/}} is employed throughout.

  \input{tables/main_results.tex}

  \paragraph{Document collection and preprocessing.}

  To achieve broad domain coverage and a large document pool, we survey several occupational classification systems\footnote{O*NET, GB/T 6565-2015, ISCED-F 2013, and LinkedIn Job Functions} and distill eight major domains as \emph{domain seeds}. A deep-search agent then retrieves semantically relevant web documents and produces a candidate list from which we verify and download the final set. The collected documents span four formats and are chunked for fine-grained retrieval by format-specific rules. Each \textbf{PDF} is converted to Markdown via \texttt{dots.ocr}~\citep{li2025dotsocrmultilingualdocumentlayout} and grouped into five-page windows that each form one chunk. Each \textbf{Markdown} file is tokenized into $8{,}000$-token sliding windows ($800$-token overlap; $7{,}200$-token stride) that each form one chunk. For \textbf{Excel} every non-empty sheet becomes one chunk represented as a compact table profile of column names, data types, summary statistics, and sample rows. Each \textbf{CSV} file is parsed as a single sheet and mapped to one chunk under the same profile. All four formats are thus normalized into plain text that a language agent can read and reason over directly, and the resulting per-domain chunks are consolidated with their metadata into a single JSON file.

  We score each chunk by an information-density heuristic that prioritizes numerics, tables, and time-series content while filtering out directory listings and standalone titles (Appendix~\ref{secapp:information-density-heuristic}), retaining the top-$100$ chunks of each domain as \emph{seed chunks} that serve as entry points for task synthesis. Finally, we use GPT-5.5 to analyze the chunks, generating summaries and tags at three hierarchical levels: \emph{chunk}, \emph{document}, and \emph{domain}. Each summary is concatenated with its tags, encoded into a dense vector, and indexed in a vector database, supporting multi-granularity evidence retrieval during task synthesis.

  \paragraph{Task synthesis.}
  To synthesize high-quality multi-hop questions at scale, an agent retrieves cross-document evidence and drafts questions grounded in it, which two further stages then refine and harden. \textit{Drafting.} Given a seed chunk, a semantic-search tool over the vector database, and a set of constraints (e.g., prohibited reasoning shortcuts, minimum hop counts, and answer-leakage criteria), the agent explores the domain corpus, identifies cross-document bridging facts, and produces a candidate task with its reference reasoning path and verification code, followed by a self-check. \textit{Refinement.} This stage enforces consistency, naturalness, and unambiguity while preserving intrinsic difficulty. The agent attempts the question to gauge answerability and leakage, checks the alignment among the question stem, reasoning chain, and reference answer, and audits properties such as cross-file coverage. Any question failing a check is rewritten. \textit{Obfuscation.} This stage removes residual leakage that survives refinement, of two kinds: \textit{lexical leakage}, where stem terms retrieve the evidence within one or two searches, and \textit{structural leakage}, where entities the solver should infer are stated outright. A suite of attack tests flags both, after which exposed terms and entity names are rewritten as business-role descriptions or equivalent expressions that preserve the original cross-file dependency structure; each rewrite is re-tested to confirm removal.

  \paragraph{Quality control.}
  We further subject the synthesized tasks to a quality-control procedure targeting difficulty, correctness, and unambiguity. Each task is first presented to \texttt{DeepSeek-V4-Pro}\footnote{All references to the \texttt{DeepSeek-V4} model family in this paper refer to its preview versions.} in a closed-book setting, and any task solvable from parametric knowledge alone is discarded. The surviving tasks are then assessed by a three-model panel (\texttt{GPT-5.5}, \texttt{DeepSeek-V4-Flash}, and \texttt{DeepSeek-V4-Pro}), and those solved correctly by all three are eliminated to guarantee sufficient difficulty. Next, \texttt{Codex} reviews each task under two conditions: conditioned solely on the query, and conditioned on the query together with the reference reasoning path from synthesis. From the two resulting trajectories, it adjudicates whether the underlying reasoning chain is valid, establishing the task's solvability and unambiguity, and only tasks passing this verification are retained. Finally, we conduct a round of human annotation grounded in both the query and its reference reasoning path, yielding a curated set of 362 tasks.

  \section{Experiments}

  \subsection{Setup}

  \paragraph{Models.}
  We evaluate GPT-5.5~\citep{openai2026gpt55systemcard}, Gemini-3.1-Pro~\citep{google2026gemini31prosystemcard}, Gemini-3.1-Flash-Lite~\citep{google2026gemini31flashlitesystemcard}, DeepSeek-V4-Flash and DeepSeek-V4-Pro~\citep{deepseek2026deepseekv4techreport}, GLM-5.1~\citep{zhipu2026glm51systemcard}, and Qwen3.5-35B-A3B and Qwen3.5-9B~\citep{qwen35blog}. All models run at temperature 1.0 with their reasoning effort set to the maximum supported. We serve the Qwen models locally using SGLang~\citep{DBLP:conf/nips/ZhengYXS0YCKSGB24}, whereas the remaining models are accessed through their respective official API providers.

  \paragraph{Agent harness.}
  We evaluate every model inside a minimal harness \texttt{mini-swe-agent}~\citep{DBLP:conf/nips/YangJWLYNP24}.\footnote{We provide additional analyses on other agent harnesses in Appendix~\ref{app:harness-ablation}.} Agents can explore the collection and run computation through bash commands, and report the final answer by printing an \texttt{<answer>}\,\dots\,\texttt{</answer>} tag, upon which the harness terminates the episode. We show the system prompt in Appendix~\ref{secapp:prompts}.

  \paragraph{Runtime environment.}
  Each task runs in an isolated E2B\footnote{\url{https://e2b.dev}} sandbox with the document collection mounted as a local directory and no internet access. The full environment specification is given in Appendix~\ref{secapp:sandbox-environment}.

  \paragraph{Budget and termination.}
  Each episode is capped at 200 interaction turns and a 3{,}600-second timeout, and terminates when the agent emits an \texttt{<answer>} tag or exhausts its turn, time, or context budget. Any episode ending without a valid \texttt{<answer>} tag is scored as incorrect. We impose no uniform context-window limit; each model operates under its own native maximum context length.

  \subsection{Main Results}

  \paragraph{\textsc{Agora} is far from saturated, and performance splits into two tiers.}
  Table~\ref{tab:main_results} reports per-domain and overall accuracy for all eight models, ordered by overall accuracy. No model exceeds $60\%$: even the strongest, Gemini-3.1-Pro, answers only $59.39\%$ of queries correctly. Since every task admits a single verifiable numeric answer solvable from the mounted collection alone, this gap does not reflect formatting artifacts but a genuine capability deficit: archive-grounded agentic document reasoning remains unsolved for current models. The eight models further split into two sharply separated tiers. A frontier tier of five occupies a $40$--$60\%$ band, while the remaining three fall well below it. The $28.73$-point gap between the tiers exceeds any gap within either. The lower tier moreover does not trail uniformly but approaches the floor domain by domain. Qwen3.5-9B scores at or below $3\%$ in five of eight domains, and Gemini-3.1-Flash-Lite at or below $7\%$ in six, leaving these smaller, lower-cost models effectively non-functional on \textsc{Agora}. We examine their failure modes in Section~\ref{sec:failure}.

  \begin{figure}[t]
    \centering
    \includegraphics[width=\linewidth]{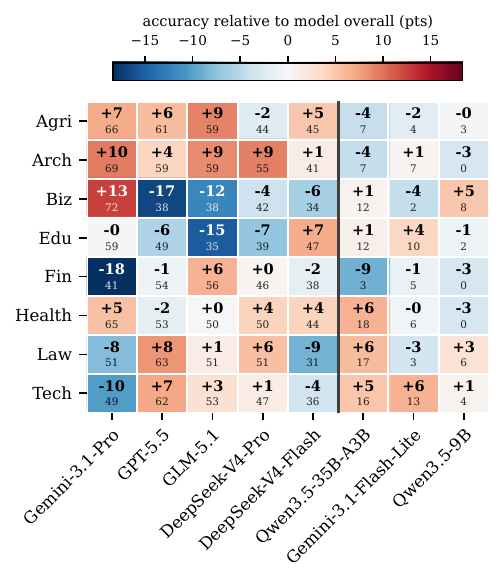}
    \caption{Per-domain accuracy relative to each model's overall accuracy on \textsc{Agora}. Each cell shows the signed residual (large) and raw accuracy (small); red is above the model's average, blue below. The vertical rule separates frontier-tier from near-floor models.}
    \label{fig:domain-variance}
  \end{figure}

  \paragraph{Per-domain accuracy varies and reorders the leaderboard.}

  Aggregate accuracy masks substantial per-domain variation in two ways. First, no model is uniformly strong: Gemini-3.1-Pro, the overall leader, still drops to $41.03\%$ on Finance, and GPT-5.5 falls to $38.00\%$ on Business. Strong aggregate performance can still mask unreliable behavior on a model's weak domains. Second, per-domain rankings diverge from the aggregate one: Gemini-3.1-Pro tops five of eight domains yet ranks only fourth on Finance, behind GLM-5.1, GPT-5.5, and DeepSeek-V4-Pro, while GPT-5.5 leads on Law and Tech. A leaderboard built on any single domain would therefore misrank these systems. These effects compound across domains, and we further analyze them in Section~\ref{sec:domain-variance}.

  Additional validation supports these findings. On a representative 40-item subset, human annotators achieve 82.5\% accuracy compared with 60.0\% for the strongest model on the same items. Expanding the retrieval scope to a flat all-document archive reduces GPT-5.5 accuracy from 54.81\% to 38.46\%, while alternative agent harnesses preserve the model ranking across all settings with complete results and reach at most 68.27\%. Appendix~\ref{app:supplementary-results} reports the full analyses.

  \section{Analysis}

  \subsection{Cross-Domain Performance Variance}
  \label{sec:domain-variance}

  \textsc{Agora}'s multi-domain design lets us ask whether agentic multi-document reasoning transfers across domains, and it does not. This is why cross-domain coverage is necessary rather than incidental to \textsc{Agora}'s design: a single-source benchmark would leave the blind spots and rank inversions below hidden. To see them, we center each model's per-domain accuracy on its overall accuracy, which isolates a domain-specific residual. As Figure~\ref{fig:domain-variance} shows, these residuals make difficulty largely a property of the model--domain pair rather than of the domain alone. Business is Gemini-3.1-Pro's strongest domain but the weakest for GPT-5.5, and the interaction is large enough to reorder the leaderboard: DeepSeek-V4-Pro trails GPT-5.5 by $8.8$ points overall yet overtakes it on Business. This imbalance does not diminish with model scale. Among frontier models the strongest is also the least even. Gemini-3.1-Pro leads overall yet has the widest \emph{spread}, the gap between its best and worst domain, at $30.97$ points. Aggregate strength and domain balance are therefore distinct axes, and a single headline number conceals the latter. The converse does not hold, however: a small spread does not imply balance. The smaller models have small spreads simply because their accuracy is near the floor across all domains.

  \subsection{Failure Modes}
  \label{sec:failure}
  \begin{figure}[t]
    \centering
    \includegraphics[width=\linewidth]{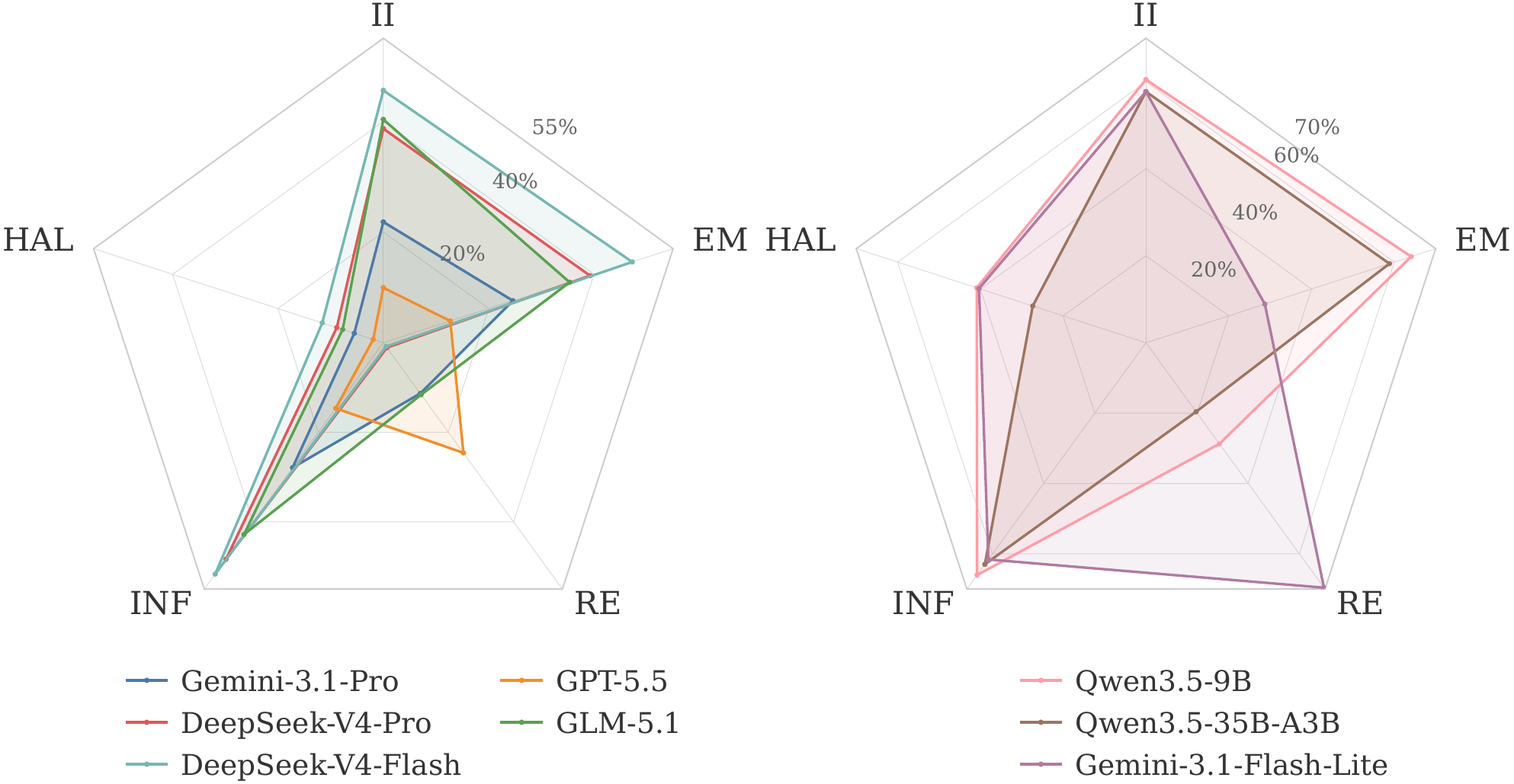}
    \caption{Failure-mode breakdown over five categories, as the share of \textsc{Agora}'s 362 questions each category is implicated in. Frontier (left, ymax\,=\,$55\%$) and lower-tier (right, ymax\,=\,$70\%$) models use different radial scales. The abbreviations are explained in Section~\ref{sec:failure}.}
    \label{fig:failure-radar}
  \end{figure}

  We annotate every wrong trace and consolidate the labels into five categories: \textit{Incomplete Inspection~(II)}, where the agent skips a required document; \textit{Evidence Misidentification~(EM)}, where it inspects the right files but extracts the wrong value; \textit{Resource Exhaustion~(RE)}, an exhausted context window, turn, time, or sandbox budget; \textit{Instruction Non-Following~(INF)}, where the agent ignores a stated requirement of the query; and \textit{Hallucination~(HAL)}, fabricated answers or forgotten earlier findings. Figure~\ref{fig:failure-radar} reveals three patterns: (i) errors across nearly all models are dominated by the three evidence-grounded categories (II, EM, INF), indicating the bottleneck lies in locating and applying evidence rather than in computation or invention; (ii) Resource Exhaustion is model-specific---GPT-5.5's top failure at $24.59\%$, near-zero for the DeepSeek-V4 family ($\leq\!1.10\%$), and catastrophic for Gemini-3.1-Flash-Lite ($69.61\%$); and (iii) Hallucination stays below $12\%$ across the frontier tier but climbs to $\sim\!40\%$ for the small models, suggesting the tier gap reflects evidence discipline rather than reasoning depth.

  \subsection{Interaction-Turn Distribution}
  \label{sec:turn-distribution}

  \begin{figure}[t]
    \centering
    \includegraphics[width=\linewidth, trim={7 3 7 5}, clip]{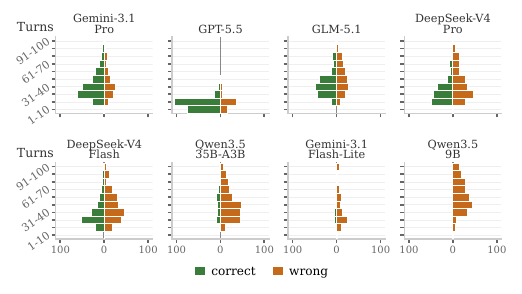}
    \caption{Distribution of interaction turns per model on \textsc{Agora}. Episodes are stratified by the turn at which the agent emits its final \texttt{<answer>} tag and partitioned by outcome; the parenthesized count denotes submitted episodes. The axis is truncated at 100, as longer episodes are rare.}
    \label{fig:turn-distribution}
  \end{figure}

  To characterize how agents allocate their exploration budget, we examine the distribution of interaction turns at which a final answer is submitted. Figure~\ref{fig:turn-distribution} stratifies each model's episodes by submission turn and partitions them by outcome. Across models, correct outcomes are concentrated at low-to-moderate turn counts, whereas episodes extending deep into the budget are predominantly incorrect. This skew suggests that prolonged trajectories more often reflect a loss of direction---repeated, unproductive exploration---than gradual convergence toward a solution and that an agent failing to resolve a task early rarely recovers by searching longer.

  \section{Conclusion}

  We presented \textsc{Agora}, an archive-grounded, cross-domain benchmark for agentic workplace document reasoning, pairing 362 verifiable numeric queries with eight domain collections of 9{,}664 documents and 372M tokens. Built by an agentic pipeline combining cross-document task synthesis, leakage-preventing obfuscation, and difficulty filtering with human verification, \textsc{Agora} jointly stresses archive-groundedness, agentic exploration, and cross-domain coverage. Evaluating eight models, we find the task far from solved: even the strongest reaches only 59.4\% accuracy, and per-domain analysis reveals systematic blind spots and rank inversions a single-source benchmark cannot surface. We hope \textsc{Agora} serves as a rigorous, reproducible testbed for the next generation of document-reasoning agents.

  \section*{Limitations}

  \textsc{Agora} spans eight professional domains distilled from official occupational classification systems—broader than prior single-source benchmarks, though not exhaustive of all workplace settings. We keep the query count (362) deliberately modest in favor of per-task quality, with every query passing multi-stage difficulty filtering, automated verification, and human annotation.

  We discard tasks solvable from parametric knowledge alone via a closed-book filtering stage, ensuring current models must genuinely reason over the mounted collection. As models scale, however, pretraining corpora may absorb more documents of this kind, eroding the guarantee; we therefore view our pipeline as a reusable instrument and hope it can refresh the benchmark with new collections as the frontier advances.

  \bibliography{custom}

  \appendix

  \section{Supplementary Experimental Results}
  \label{app:supplementary-results}

  \subsection{Difficulty-Filtering Audit}
  \label{app:filtering-audit}

  To test whether discarding tasks solved by all three panel models biases the retained benchmark against them, we ran the two strongest non-panel models on a random sample of discarded tasks (up to 24 per domain, subject to availability; domains with no discarded tasks are omitted; 102 tasks in total), under the same harness and sandbox as the main experiments.

  \input{tables/filtering_audit.tex}

  As shown in Table~\ref{tab:discarded-tasks}, both non-panel models solve 92-95\% of the discarded tasks, versus about 59.39\% and 50.00\% on the retained benchmark (Table~\ref{tab:main_results}). This suggests the discarded tasks tend to be easy for frontier models in general rather than for the panel specifically, indicating our panel is reasonably representative and does not appear to favor particular models.

  \subsection{Successful-Trajectory Audit}
  \label{app:success-audit}

  We manually inspected all 413 correct trajectories produced by the two strongest models for two types of shortcuts: (i) directly reading the answer and (ii) reaching it without the intended reasoning; we found no evidence of either.

  \begin{itemize}
      \item No answer leakage. We scanned all 724 command traces of the two strongest models (GPT-5.5, Gemini-3.1-Pro). No correct trajectory reads any gold-answer, evaluation-metadata, or other-model-result file; the rare episodes that scan the whole sandbox end in wrong answers, so no such access yields a correct score. A benchmark-wide exploitable shortcut is also inconsistent with the outcome structure: 56 tasks are solved by no model and 172 by at most two models, even though any model could invoke such a shortcut.
      \item Correct answers come from genuine reasoning, not a lookup. In the correct trajectories we inspected, wins reflect scope control, structured parsing, and script-based recomputation. For example, on q0020 (solved only by GPT-5.5), the correct run writes a stateful HTML-table parser that tracks document type, sector, and inherited headers, drops incomplete years, and computes the growth rate in code. Models that string-matched instead mis-scoped the table or applied factor$\times$100 rather than (factor-1)$\times$100, returning ~151\% instead of ~51\%. Consistent with this, correct runs are not one-shot: GPT-5.5's have a median of 12 tool calls (2 of 198 within 5) and Gemini-3.1-Pro's a median of 34 (none within 5).
  \end{itemize}
  
  In addition, we find that the scores are not carried by trivial questions. Excluding the 26 consensus-easy tasks (solved by $\geq$ 6/8 models) leaves ranking and gaps intact (Gemini-3.1-Pro 59.39$\to$56.25, GPT-5.5 54.70$\to$51.19, GLM-5.1 50.00$\to$46.13, DeepSeek-V4-Pro 45.86$\to$41.67), each model dropping only ~3–4 pts.

  \subsection{Human Baseline and Gold-Answer Audit}
  \label{app:human-audit}
  
  We sampled five questions per domain, 40 in total, drawn randomly under a constraint computed only from pre-existing model results: each model's subset accuracy stays close to Table~\ref{tab:main_results} and the ranking is preserved, so the slice mirrors full-benchmark difficulty and cannot favor the human side. Annotators worked in a VS Code\footnote{\url{https://github.com/microsoft/vscode}} editor with only the paired document collection and editor's text search backed by ripgrep\footnote{\url{https://github.com/BurntSushi/ripgrep}}, the same tool available in the agent sandbox, blind to gold answers and reference paths, and never on items they helped create.

  \input{tables/human_baseline.tex}

  As shown in Table~\ref{tab:human-baseline}, annotators achieve 82.5\% accuracy, compared with 60.0\% for the strongest model on the same subset. The annotators are graduate researchers, not certified professionals. That means the questions are solvable through careful navigation, cross-file reconciliation, and arithmetic rather than specialist knowledge, which is exactly the capability \textsc{Agora} measures. After completing the evaluation, annotators re-derived all 40 answers using the reference reasoning paths and confirmed all 40 gold answers, including those initially answered incorrectly.

  \subsection{Retrieval-Scope Ablation}
  \label{app:retrieval-scope}

  We evaluated three document-scope settings on a representative 104-task subset containing 13 questions per domain. For each evaluated model, accuracy on this subset is within approximately one percentage point of its full-benchmark accuracy in Table~\ref{tab:main_results}, and the model ranking is preserved. The question, evidence, and gold answer remain unchanged across settings.

  \input{tables/retriebal_scope.tex}

  Pooling all documents without domain labels drops accuracy sharply (GPT-5.5 54.81 → 38.46), since the evidence and answer are unchanged, but the search space grows—isolating retrieval difficulty. We keep paired-domain as the primary setting because it is already unsaturated (<60\%) and its scoped evidence set preserves deterministic verification.

  \subsection{Agent-Harness Ablation}
  \label{app:harness-ablation}

  We reran the 104-task subset introduced in Appendix~\ref{app:retrieval-scope} under mini-swe-agent~\citep{DBLP:conf/nips/YangJWLYNP24}, OpenHands~\citep{DBLP:conf/iclr/0001LSXTZPSLSTL25}, and OpenClaw\footnote{\url{https://github.com/openclaw/openclaw}}.

  \input{tables/harness_ablation.tex}

  As shown in Table~\ref{tab:harness-ablation}, (i) the best overall score on this 104-task subset is 68.27\% (OpenHands with GPT-5.5), indicating that the benchmark remains unsaturated under the evaluated harnesses; (ii) the model ranking is stable in both harnesses for which all three models completed: GPT-5.5 $>$ DeepSeek-V4-Pro $\gg$ Qwen3.5-35B-A3B; GPT-5.5 also remains ahead of DeepSeek-V4-Pro under OpenClaw; (iii) richer harnesses do not improve all models uniformly.

  \subsection{Preprocessing and OCR Analysis}
  \label{app:ocr-analysis}

  Only PDF files pass through OCR; Markdown, Excel, and CSV files are processed using deterministic parsers. Of the 764,550 PDF pages, 31,300 failed to parse, corresponding to a failure rate of 4.094\%. Failed pages produce no Markdown and are excluded from both task synthesis and task solving, and therefore cannot serve as evidence for a released task.

  For successfully processed pages, task construction and evaluation operate on byte-identical post-OCR Markdown. Gold-answer correctness with respect to the processed corpus was manually checked during construction for all 362 tasks and independently reconfirmed for all 40 items in the human audit.

  \section{Sandbox Environment}
  \label{secapp:sandbox-environment}

  All \textsc{Agora} tasks run inside an isolated E2B sandbox with 2~vCPUs, 4~GB of memory, no network access, and a 3{,}600-second timeout. The sandbox is built from a fixed Docker template that extends the official \texttt{e2bdev/base} image with a Python~3 interpreter, a few command-line utilities (\texttt{jq}, \texttt{ripgrep}, \texttt{fd}, \texttt{tree}, \texttt{less}), and a pinned set of scientific-computing and document-parsing libraries: \texttt{pandas}, \texttt{numpy}, \texttt{scipy}, \texttt{openpyxl}, \texttt{lxml}, \texttt{beautifulsoup4}, \texttt{pyyaml}, \texttt{pymupdf}, \texttt{pdfplumber}, and \texttt{tabulate}. These cover the tabular computation and file parsing that \textsc{Agora} tasks require, without any network access for installing further dependencies. The working directory is \texttt{/workspace}, with the document collection at \texttt{/workspace/documents} and \texttt{/workspace/run} as scratch space. All package versions are pinned in the template so the environment is stable across evaluation runs.

  \section{Heuristic for Information-Density Scoring}
  \label{secapp:information-density-heuristic}

  We rank candidate chunks for downstream task synthesis with a lightweight additive heuristic, applied during preprocessing to direct the more expensive agentic synthesis toward chunks whose content can plausibly support multi-hop computation. Because synthesis is invoked once per seed chunk and incurs a full agent trajectory, even a coarse prefilter materially reduces wasted budget; we therefore favor an inexpensive rule-based score over an LLM-judged one. The signals we accumulate track three properties typical of evidence in workplace reasoning tasks: numeric density, tabular structure, and temporal extent.

  \paragraph{Scoring formula.}

  For a chunk with raw text $c$ and metadata $m$, we sum the contributions below, each clamped to its listed cap so that no single signal can dominate:
  \begin{itemize}
    \setlength{\itemsep}{1pt}

    \item \textit{Numeric density:} $5 \cdot 1000 \cdot |N(c)| / |c|$, capped at $30$, where $N(c)$ is the set of numeric tokens in $c$. This rewards chunks dense in figures such as financial values, counts, or measurements.

    \item \textit{Length:} $\mathrm{tokens}(c) / 50$, capped at $20$. Slightly larger chunks have more room to host the cross-row computation that multi-hop questions require.

    \item For \texttt{Excel} and \texttt{CSV} chunks we further add: $\mathrm{rows}/100$ (cap $15$), $\mathrm{cols}$ (cap $10$), $2 \cdot |\textrm{numeric columns}|$, $8$ if any time column is present, and $2 \cdot (\mathrm{year}_{\max}- \mathrm{year}_{\min})$ (cap $10$). Together they reward wide, numerically typed, time-spanning sheets.

    \item For \texttt{Markdown} chunks we add $10$ if a table is present and $3$ if a list is present; for \texttt{PDF} chunks we add $12$ when the OCR'd output contains a table. Tabular structure is a strong cue that the chunk admits cell-level lookup.
  \end{itemize}

  \paragraph{Hard filters.}

  Two conditions force a sentinel score that excludes the chunk regardless of its other signals. \emph{(i) Token floor.} Chunks with fewer than $20$ tokens---typically standalone titles, section headers, or stray footer fragments---are too sparse to bridge across documents and are dropped. \emph{(ii) Directory listings.} Chunks whose leading $200$ characters match a table-of-contents pattern (\texttt{table of contents}, \texttt{contents}, \texttt{index}) are structural artifacts of their parent document and contribute no extractable evidence.

  \paragraph{Selection.}

  For each domain we score every chunk, discard those falling below a minimum-score floor of $10$, sort the remainder by score, and retain the top-$100$ as seed chunks for task synthesis (Section~\ref{sec:construction}). The floor prunes a long tail of low-information chunks; the per-domain cap bounds the synthesis budget and prevents richer corpora from monopolizing the seed pool while ensuring sparser ones still contribute. We deliberately keep the heuristic mechanical: it is a router for downstream agentic synthesis, not itself a quality judgment, and the subsequent refinement, obfuscation, and quality-control stages remain responsible for filtering tasks that survive selection but fail later checks.

  \section{Prompts}
  \label{secapp:prompts}
  \input{prompts}

  \section{Responsible NLP Statements}
  All models evaluated in this work are used strictly in accordance with their respective original licenses and intended-use terms. We confirm that our use of existing artifacts is consistent with their intended use. The dataset we introduce is released strictly for academic research purposes; it is a derivative of data accessed for research purposes and must not be used in any commercial or non-research context, in line with the original access conditions. The data used in this work is derived from publicly available sources, which do not contain private personally identifying information.

  We discuss the potential risks of this work. The document collections in \textsc{Agora} are compiled from publicly available workplace files, and we release them strictly for research purposes; users should respect the terms of the original sources and avoid uses beyond academic evaluation. More broadly, \textsc{Agora} is intended to drive progress in agents that can autonomously explore and reason over large document archives, and advances in such AI capabilities may carry wider societal implications that warrant ongoing attention. Finally, as pretraining corpora continue to grow, benchmark documents may gradually be absorbed into training data, weakening the closed-book filtering safeguard against knowledge leakage and causing evaluation results to drift over time.

  We acknowledge the use of large language models, specifically Claude Opus 4.7 and GPT-5.5, as writing aids during the preparation of this manuscript. Their use was limited to language polishing, grammar correction, literature search, code completion, and improving the clarity of presentation.

  \section{Task Examples}
  \label{secapp:task_example}

  \input{task_example}
\end{document}

%% file: tables/benchmark_comparison.tex
\begin{table*}
    [t]
    \centering
    \footnotesize
    \setlength{\tabcolsep}{6pt}
    \renewcommand{\arraystretch}{1.15}
    \begin{tabular}{l l r c c c}
        \toprule \textbf{Benchmark}                            & \textbf{Corpus}         & \textbf{\#Test} & \textbf{Multi-file} & \textbf{Archive-grounded} & \textbf{Agentic} \\
        \midrule HotpotQA~\citep{DBLP:conf/emnlp/Yang0ZBCSM18} & Wikipedia               & 7{,}405         & \cmark              & \cmark                    & \xmark           \\
        MuSiQue~\citep{DBLP:journals/tacl/TrivediBKS22}        & Wikipedia               & 2{,}459         & \cmark              & \cmark                    & \xmark           \\
        TAT-QA~\citep{DBLP:conf/acl/ZhuLHWZLFC20}              & Financial reports       & 1{,}669         & \xmark              & \cmark                    & \xmark           \\
        FinanceBench~\citep{DBLP:journals/corr/abs-2311-11944} & U.S. SEC filings        & 150             & \pmark              & \cmark                    & \xmark           \\
        FRAMES~\citep{DBLP:conf/naacl/KrishnaKMSSUF25}         & Wikipedia               & 824             & \cmark              & \cmark                    & \xmark           \\
        OfficeQA Pro~\citep{DBLP:journals/corr/abs-2603-08655} & U.S. Treasury bulletins & 133             & \cmark              & \cmark                    & \cmark           \\
        GAIA~\citep{DBLP:conf/iclr/MialonF0LS24}               & Open Web                & 165             & --                  & \xmark                    & \cmark           \\
        BrowseComp~\citep{DBLP:journals/corr/abs-2504-12516}   & Open Web                & 1{,}266         & --                  & \xmark                    & \cmark           \\
        \midrule \rowcolor{gray!10} \textsc{Agora}~(Ours)      & Multi-domain office     & 362             & \cmark              & \cmark                    & \cmark           \\
        \bottomrule
    \end{tabular}
    \caption{ Comparison with representative multi-hop QA, document QA, RAG, and agent benchmarks. \textbf{Multi-file}: whether answering a question may require evidence from multiple files. \textbf{Archive-grounded}: whether reasoning is restricted to a fixed corpus rather than live web search or an open environment. \textbf{Agentic}: whether answering requires active file navigation and code execution. Marks: \cmark{} = yes, \pmark{} = partial, \xmark{} = no, -- = not applicable. }
    \label{tab:benchmark_comparison}
\end{table*}

%% file: tables/benchmark_composition.tex
\begin{table}[t]
    \centering
    \footnotesize
    \begin{tabular}{lrrrr}
        \toprule \textbf{Domain}  & \textbf{\# Docs}     & \textbf{\# Tokens} & \textbf{\# Questions} \\
        \midrule Agri             & 118                  & 38M                & 71                    \\
        Arch                      & 1{,}005              & 48M                & 29                    \\
        Biz                       & 358                  & 22M                & 50                    \\
        Edu                       & 152                  & 29M                & 49                    \\
        Fin                       & 5{,}966              & 117M               & 39                    \\
        Health                    & 1{,}233              & 41M                & 34                    \\
        Law                       & 285                  & 30M                & 35                    \\
        Tech                      & 547                  & 47M                & 55                    \\
        \midrule \textbf{Overall} & \textbf{9{,}664}     & \textbf{372M}      & \textbf{362}          \\
        \bottomrule
    \end{tabular}
    \caption{Composition of the dataset. Domains are abbreviated as Agri (Agriculture, Resources \& Energy), Arch (Architecture, Construction, Real Estate \& Facilities), Biz (Business, Management, Marketing \& Sales), Edu (Education, Science \& Academia), Fin (Finance \& Economics), Health (Healthcare \& Medicine), Law, and Tech (Technology, Software \& Manufacturing).}
    \label{tab:composition}
\end{table}

%% file: tables/main_results.tex
\begin{table*}
    [t]
    \centering
    \setlength{\tabcolsep}{5pt}
    \begin{tabular}{lccccccccc}
        \toprule                          & \multicolumn{8}{c}{\textbf{By Domain}} &                    \\
        \cmidrule(lr){2-9} \textbf{Model} & Agri                                   & Arch              & Biz               & Edu               & Fin               & Health            & Law               & Tech              & \textbf{Overall}  \\
        \midrule \texttt{Gemini-3.1-Pro}  & \textbf{66.20}                         & \textbf{68.97}    & \textbf{72.00}    & \textbf{59.18}    & 41.03             & \textbf{64.71}    & \underline{51.43} & 49.09             & \textbf{59.39}    \\
        \texttt{GPT-5.5}                  & \underline{60.56}                      & \underline{58.62} & 38.00             & \underline{48.98} & \underline{53.85} & \underline{52.94} & \textbf{62.86}    & \textbf{61.82}    & \underline{54.70} \\
        \texttt{GLM-5.1}                  & 59.15                                  & \underline{58.62} & 38.00             & 34.69             & \textbf{56.41}    & 50.00             & \underline{51.43} & \underline{52.73} & 50.00             \\
        \texttt{DeepSeek-V4-Pro}          & 43.66                                  & 55.17             & \underline{42.00} & 38.78             & 46.15             & 50.00             & \underline{51.43} & 47.27             & 45.86             \\
        \texttt{DeepSeek-V4-Flash}        & 45.07                                  & 41.38             & 34.00             & 46.94             & 38.46             & 44.12             & 31.43             & 36.36             & 40.06             \\
        \texttt{Qwen3.5-35B-A3B}          & 7.04                                   & 6.90              & 12.00             & 12.24             & 2.56              & 17.65             & 17.14             & 16.36             & 11.33             \\
        \texttt{Gemini-3.1-Flash-Lite}    & 4.23                                   & 6.90              & 2.00              & 10.20             & 5.13              & 5.88              & 2.86              & 12.73             & 6.35              \\
        \texttt{Qwen3.5-9B}               & 2.82                                   & 0.00              & 8.00              & 2.04              & 0.00              & 0.00              & 5.71              & 3.64              & 3.04              \\
        \bottomrule
    \end{tabular}
    \caption{Per-domain and overall accuracy (\%) on \textsc{Agora}. Models are ordered by overall accuracy. The best result in each column is in \textbf{bold} and the second best is \underline{underlined}. Domain abbreviations follow Table~\ref{tab:composition}.}
    \label{tab:main_results}
\end{table*}

%% file: tables/filtering_audit.tex
\begin{table}[htbp]
    \centering
    \small \resizebox{\columnwidth}{!}{%
    \begin{tabular}{lrrrrrr}
        \toprule \textbf{Model}  & \textbf{Agri.} & \textbf{Arch.} & \textbf{Biz.} & \textbf{Edu.} & \textbf{Fin.} & \textbf{Overall} \\
        \midrule Number of tasks & 24             & 24             & 24            & 24            & 6             & 102              \\
        Gemini-3.1-Pro           & 95.8           & 91.7           & 95.8          & 87.5          & 83.3          & 92.2             \\
        GLM-5.1                  & 91.7           & 91.7           & 100.0         & 95.8          & 100.0         & 95.1             \\
        \bottomrule
    \end{tabular}}
    \caption{Accuracy (\%) of non-panel models on discarded tasks.}
    \label{tab:discarded-tasks}
\end{table}

%% file: tables/human_baseline.tex
\begin{table*}
    [htbp]
    \centering
    \small 
    % \resizebox{\textwidth}{!}{%
    \begin{tabular}{lrrrrrrrrr}
        \toprule  & \textbf{Agri.} & \textbf{Arch.} & \textbf{Biz.} & \textbf{Edu.} & \textbf{Fin.} & \textbf{Health} & \textbf{Law} & \textbf{Tech.} & \textbf{Overall} \\
        \midrule \textbf{Human}           & 80             & 80             & 80            & 100           & 100           & 60              & 100          & 60             & 82.5             \\
        Gemini-3.1-Pro           & 60             & 80             & 40            & 60            & 40            & 60              & 80           & 60             & 60.0             \\
        GPT-5.5                  & 60             & 60             & 20            & 80            & 20            & 60              & 100          & 40             & 55.0             \\
        GLM-5.1                  & 60             & 60             & 0             & 40            & 60            & 60              & 80           & 40             & 50.0             \\
        DeepSeek-V4-Pro          & 40             & 80             & 0             & 40            & 60            & 40              & 60           & 40             & 45.0             \\
        DeepSeek-V4-Flash        & 40             & 40             & 0             & 60            & 40            & 60              & 40           & 40             & 40.0             \\
        Qwen3.5-35B-A3B          & 0              & 0              & 20            & 20            & 0             & 0               & 40           & 20             & 12.5             \\
        Gemini-3.1-Flash-Lite    & 0              & 20             & 0             & 0             & 0             & 0               & 20           & 20             & 7.5              \\
        Qwen3.5-9B               & 0              & 0              & 0             & 0             & 0             & 0               & 0            & 20             & 2.5              \\
        \bottomrule
    \end{tabular}
    % }
    \caption{Accuracy (\%) on the 40-item human-evaluation subset.}
    \label{tab:human-baseline}
\end{table*}

%% file: tables/retriebal_scope.tex
\begin{table*}
    [t]
    \centering
    \scriptsize \resizebox{\textwidth}{!}{%
    \begin{tabular}{llrrrrrrrrr}
        \toprule \textbf{Document scope} & \textbf{Model}  & \textbf{Agri.} & \textbf{Arch.} & \textbf{Biz.} & \textbf{Edu.} & \textbf{Fin.} & \textbf{Health} & \textbf{Law} & \textbf{Tech.} & \textbf{Overall} \\
        \midrule Paired domain           & GPT-5.5         & 61.54          & 61.54          & 38.46         & 46.15         & 53.85         & 53.85           & 61.54        & 61.54          & 54.81            \\
        Paired domain                    & DeepSeek-V4-Pro & 46.15          & 53.85          & 38.46         & 38.46         & 46.15         & 53.85           & 46.15        & 46.15          & 46.15            \\
        Paired domain                    & Qwen3.5-35B-A3B & 7.69           & 7.69           & 15.38         & 15.38         & 0.00          & 15.38           & 15.38        & 15.38          & 11.54            \\
        \midrule All domains, foldered   & GPT-5.5         & 61.54          & 38.46          & 53.85         & 38.46         & 61.54         & 46.15           & 76.92        & 76.92          & 56.73            \\
        All domains, foldered            & DeepSeek-V4-Pro & 53.85          & 53.85          & 38.46         & 38.46         & 15.38         & 38.46           & 38.46        & 46.15          & 40.38            \\
        All domains, foldered            & Qwen3.5-35B-A3B & 7.69           & 7.69           & 7.69          & 15.38         & 0.00          & 0.00            & 7.69         & 7.69           & 6.73             \\
        \midrule All documents, flat     & GPT-5.5         & 53.85          & 30.77          & 23.08         & 23.08         & 53.85         & 38.46           & 46.15        & 38.46          & 38.46            \\
        All documents, flat              & DeepSeek-V4-Pro & 30.77          & 53.85          & 30.77         & 30.77         & 46.15         & 38.46           & 23.08        & 46.15          & 37.50            \\
        All documents, flat              & Qwen3.5-35B-A3B & 0.00           & 0.00           & 7.69          & 7.69          & 7.69          & 7.69            & 0.00         & 0.00           & 3.85             \\
        \bottomrule
    \end{tabular}}
    \caption{Accuracy (\%) under different document-scope settings.}
    \label{tab:retrieval-scope}
\end{table*}

%% file: tables/harness_ablation.tex
\begin{table*}
    [t]
    \centering
    \scriptsize \resizebox{\textwidth}{!}{%
    \begin{tabular}{llrrrrrrrrr}
        \toprule \textbf{Harness} & \textbf{Model}  & \textbf{Agri.} & \textbf{Arch.} & \textbf{Biz.} & \textbf{Edu.} & \textbf{Fin.} & \textbf{Health} & \textbf{Law} & \textbf{Tech.} & \textbf{Overall} \\
        \midrule mini-swe-agent   & GPT-5.5         & 61.54          & 61.54          & 38.46         & 46.15         & 53.85         & 53.85           & 61.54        & 61.54          & 54.81            \\
        mini-swe-agent            & DeepSeek-V4-Pro & 46.15          & 53.85          & 38.46         & 38.46         & 46.15         & 53.85           & 46.15        & 46.15          & 46.15            \\
        mini-swe-agent            & Qwen3.5-35B-A3B & 7.69           & 7.69           & 15.38         & 15.38         & 0.00          & 15.38           & 15.38        & 15.38          & 11.54            \\
        \midrule OpenHands        & GPT-5.5         & 61.54          & 69.23          & 76.92         & 76.92         & 69.23         & 53.85           & 76.92        & 61.54          & 68.27            \\
        OpenHands                 & DeepSeek-V4-Pro & 46.15          & 69.23          & 46.15         & 30.77         & 61.54         & 53.85           & 46.15        & 61.54          & 51.92            \\
        OpenHands                 & Qwen3.5-35B-A3B & 0.00           & 0.00           & 7.69          & 0.00          & 0.00          & 15.38           & 0.00         & 0.00           & 2.88             \\
        \midrule OpenClaw         & GPT-5.5         & 69.23          & 53.85          & 53.85         & 61.54         & 76.92         & 61.54           & 46.15        & 61.54          & 60.58            \\
        OpenClaw                  & DeepSeek-V4-Pro & 46.15          & 46.15          & 30.77         & 38.46         & 53.85         & 30.77           & 46.15        & 38.46          & 41.35            \\
        \bottomrule
    \end{tabular}}
    \caption{Accuracy (\%) under different agent harnesses. Qwen3.5-35B-A3B did not complete reliably under OpenClaw within the available time budget.}
    \label{tab:harness-ablation}
\end{table*}

%% file: prompts.tex
\newtcblisting{promptbox}[1][]{
  colback=gray!5, colframe=gray!50,
  fonttitle=\bfseries,
  breakable,
  listing only,
  listing options={
    basicstyle=\ttfamily\small,
    breaklines=true,
    breakindent=0pt,
    columns=fullflexible,
    keepspaces=true,
  },
  #1
}

\begin{promptbox}[title=System Prompt for Evaluation]
You are a helpful assistant working inside a Linux sandbox. You have only bash, with the usual tools available (python3, jq, grep/ripgrep, sed, awk, find, head, tail, cat, fd). Reference documents are at /workspace/documents/{{domain}}/.

Every reply must include a bash tool call.

When you are ready to answer, submit the answer by running ONE bash command whose stdout is exactly two lines, in this order:
COMPLETE_TASK_AND_SUBMIT_FINAL_OUTPUT
<answer>YOUR FINAL ANSWER</answer>

A valid final command example is:
printf 'COMPLETE_TASK_AND_SUBMIT_FINAL_OUTPUT\n<answer>42</answer>\n'
\end{promptbox}

\begin{promptbox}[title=Prompt for Chunk-Level Summary]
You are a data summarization assistant. Generate a concise summary of the provided data chunk. Include: main topic, key entities, time range (if any), data type, and notable features. Respond in the same language as the input data.
At the end of your response, output 3-8 keyword tags in this exact XML format:
<tags>tag1, tag2, tag3</tags>
The tags line MUST be present and use the <tags></tags> wrapper.
\end{promptbox}

\begin{promptbox}[title=Prompt for Task Reviewer]
# Your Role

You are an impartial and unbiased question quality reviewer. Your task is to strictly and fairly review whether a given question, reference answer, reasoning process, and provided_files constitute a reliable, deliverable task, and to perform a consistency check on these four parts.

# What You Must Know

## Question

{question}

## Reference Answer

{expected_answer}

## Reasoning Process

{reasoning}

## provided_files

{reference_files}

## Solving Workflow: Launch Two Subagents

You do not solve the problem yourself. Instead, you must launch two subagents, have each of them complete a solving task, and only after both have finished, observe their solving trajectories and conduct your review based on them.

1. **Subagent A (Blind Solve)**: Give it only the question statement, without any reference information (no reference answer, no reasoning process; provided_files may be supplied as needed), and have it solve independently.
2. **Subagent B (Solve with Reasoning Chain)**: Give it the question statement together with the reference reasoning chain (reasoning), but do **not** provide the reference answer, and have it solve by following the reference reasoning.

Requirements:
- Both subagents should deliberately think about the reasonableness of the question during their solving process, and separately record the files each of them has accessed.
- Instruct both subagents not to blindly read all the data, to avoid causing information collapse.
- Only after both subagents have completed their tasks, collect and observe their full trajectories (solving steps, files accessed, conclusions reached).
- Note: You are a strict, impartial reviewer, not a solver focused on solving the problem. The subagents' solving is only to provide information for your subsequent judgment, so no cheating methods (such as copying the answer) should be used.

## Review Requirements

Based on and comparing the solving trajectories of the two subagents and the information gathered, you need to make the following judgments:

1. Determine whether the question is clear and answerable, and whether the answer contains errors.
2. Whether the question has ambiguity or multiple reasonable answers.
3. Determine whether the reasoning process is logically valid, whether there are calculation errors, citation errors, or insufficient evidence, whether it can solve the question when the question is unambiguous, and judge whether each hop is necessary.
4. Based on the files used by the two subagents respectively, determine whether provided_files has redundancy (no help to the answer), and provide correct file statistics.
5. Determine whether key facts can be found in provided_files, and whether it relies on external information not provided.
6. Determine whether this QA sample is suitable for delivery (extremely important).
7. Finally, you must give a binary is_usable label: questions that can be directly delivered without affecting correctness are marked true; questions with problems such as answer errors, insufficient evidence, serious ambiguity, or reasoning that cannot support the answer are marked false.

Only return valid JSON, with the schema as follows:

{
  "is_usable": true,
  "quality_score": 0, // 0~5, integer
  "ambiguity": "none|minor|major",
  "answer_status": "correct|minor_issue|wrong|unclear",
  "reasoning_status": "valid|minor_issue|invalid|unclear",
  "evidence_status": "sufficient|partial|insufficient|unclear",
  "usable_files": ["string"],
  "redundant_files": ["string"],
  "missing_files": ["string"],
  "issues": ["string"],
  "review_summary": "string"
}
\end{promptbox}

%% file: task_example.tex
\lstdefinestyle{taskstyle}{
  basicstyle=\ttfamily\small,
  breaklines=true,
  breakindent=0pt,
  columns=fullflexible,
  keepspaces=true,
}

\newtcolorbox{taskbox}[1][]{
  colback=gray!5, colframe=gray!50,
  colbacklower=blue!5,
  fonttitle=\bfseries,
  breakable,
  #1
}

\begin{taskbox}[title={Agriculture, Resources \& Energy}]
\begin{lstlisting}[style=taskstyle]
A portfolio analyst is tying together Germany's renewable-support statistics with the ministry material on sectoral economic effects. In the newest regulator observation in the collection, work within the electricity-support block at the technology-row level: keep rows whose later-column growth in eligible delivered output is below the total row while the same row's later-column growth in the companion support-cash table is above the total row. From that reduced set, use the newest short ministry brief to select the row that is presented as the technology-specific capital-buildout exception, because its latest plant-building spend is described as having exceeded its own earlier high point.\n\nFor that selected technology, compare two arithmetic means. On the regulator side, use every successive observation in the collection that reports this eligible-output support table; treat each table's later-year column as that observation and its immediately preceding column in the same table as the base. On the ministry side, form the comparable plant-building investment series from exactly three ministry documents, in order: the comprehensive annual numbers compendium for the first investment year, the subsequent English-language bridge note for the middle investment year, and the newer German short brief for the latest investment year.\n\nCompute all rates from table cell values, not rounded percentage displays or prose, and put physical-output and currency scales on consistent units before forming rates. Report the regulator-side mean minus the ministry-side mean as signed percentage points rounded to two decimals, formatted as `+x.xx pp` or `-x.xx pp`.
\end{lstlisting}
\tcblower
\textbf{Answer:} -18.03 pp
\end{taskbox}

\begin{taskbox}[title={Architecture, Construction, Real Estate \& Facilities}]
\begin{lstlisting}[style=taskstyle]
An international housing dashboard needs one multiplier. For the Japan-side numerator, first use the survey report's own definitions to determine the full set of dwelling-related categories. For each category, go to its repeated cross-year table and choose the respondent-choice item that records where the respondent found the relevant dwelling or contractor; do not use the nearby item about general digital activity when it lacks the full early-to-late series. Within the chosen item, use the option for the web route, take values at the broadest geography that the survey design makes available for that category, align all resulting series to their shared local-era fiscal-year window, and keep the largest compound annual change. For the England-side denominator, use the valuation component report for a national housing survey: take the opening headline percent change for the arithmetic-average dwelling value between its two valuation reference dates, annualized by exact elapsed days using a 365.2425-day year. What is the numerator divided by the denominator? Return the quotient rounded to three decimal places, followed by `x`. Format: `<value with three decimal places>x`.
\end{lstlisting}
\tcblower
\textbf{Answer:} 5.517x
\end{taskbox}

\begin{taskbox}[title={Business, Management, Marketing \& Sales}]
\begin{lstlisting}[style=taskstyle]
An Italian connectivity retailer wants a single launch-pricing premium that compares a present-day access-and-demand basket with the older wholesale access signal that a retrospective infrastructure-investment study treats as requiring entrants to commit assets inside the incumbent's local exchange footprint, then scales the gap by the size of the top-level requirement catalogue in the card-acceptance security rulebook used by sales environments. Use only the source documents' own endpoints and displayed percentage values. The historical side is the compound annual growth of that study's retained proxy across its full observation window; any present-day component that spans a multi-year sequence uses the same compound annual-growth convention. Build the present-day side as an equal-weight average of four roles: the household reach measure for the end-to-premises fixed-access form that the small-firm white paper distinguishes from the cabinet-plus-copper alternative, carried from that paper's first displayed national household point to the later annual-report value for the same public benchmark; the matching reach measure for the address-resolved small-firm sample, carried from the white paper's research-sample result to the annual report's later update for the same firm segment and computed as simple relative growth; the fixed-service usage-volume sequence in the statistics annex, using every year displayed there; and the relative uplift between the consumer-study demand signal for people already in the top current-speed class and the business-study demand signal for the current-speed class that the business report identifies as the one most inclined to spend more. Report the present-day-minus-historical premium in percentage points per top-level requirement entry, rounded to three decimals. Answer format: `x.xxx pp/requirement`.
\end{lstlisting}
\tcblower
\textbf{Answer:} 2.463 pp/requirement
\end{taskbox}

\begin{taskbox}[title={Education, Science \& Academia}]
\begin{lstlisting}[style=taskstyle]
An analyst is studying how Chinese universities turned research results into cash through external transfer deals. First, use the recurring English executive-summary series from the national policy-research institute outside China in the corpus: take the consecutive releases in which the country that is the subject of each summary sits just behind the two larger national systems on both the money-scale and workforce-scale front-door capacity measures, and stop with the release where that country's authorship-apportioned bibliometric-output rank no longer matches the repeated position seen in the immediately preceding releases. For the Chinese higher-education S&T statistical books whose release years fall in that release span, retain only those editions whose prefatory note says the book is built from the normal province-administered annual reporting process for the regular higher-education S&T population, rather than from a special nationwide inventory of research-development resources. Respect each retained book's own institutional and territorial scope. From the national aggregate row in the broad activity-overview table, take the cash received during the year for university result-transfer deals and divide it by the count of such deals made in that year. Let beta be the OLS slope of the natural log of that cash-per-deal series on the underlying statistical year recorded in the prefatory note. Report exp(beta)-1 as one signed percentage rounded to three decimals, using the template [sign]x.xxx%.
\end{lstlisting}
\tcblower
\textbf{Answer:} -9.710\%
\end{taskbox}

\begin{taskbox}[title=Finance \& Economics]
\begin{lstlisting}[style=taskstyle]
Within the source, first identify the holdings review that ranks a primary-production stock basket by the money committed by stock-heavy fund products and says that, inside its highest-ranked basket, the largest representation remains an upstream animal-protein group. For the companies in that role-defined group whose standalone company note for the same reporting season both shows the same visible publication date as the holdings review and provides a full forward sequence for profit attributable to the parent, compute each firm's CAGR from the sequence's first forecast year to its last. Treat the different RMB magnitude labels used across narrative and tables as the same scale before calculating. What is the cross-sectional median of those CAGRs? Return one percentage rounded to three decimals, in the format `x.xxx%`.
\end{lstlisting}
\tcblower
\textbf{Answer:} 23.687\%
\end{taskbox}

\begin{taskbox}[title=Healthcare \& Medicine]
\begin{lstlisting}[style=taskstyle]
A public health analyst is comparing one admissions-side category across several releases from the same substance-treatment admissions data family. First identify the category from the older national report's special-interest section: it is the primary-substance group used to discuss planned medication support for opioid treatment, in a context that distinguishes person-entry admissions data from a facility-day client measure used elsewhere. Then gather the annual counts for that same group from the older report's multi-year primary-substance count table, from the frequency tables in the two separate single-year admission-file metadata documents, and from the later report's admissions-side comparison of the leading primary-substance groups. Using only those observations, take the largest count as the base and the chronologically latest count as the endpoint. What compound annual rate of change do they imply? Round the signed percentage to three decimals.\n\nAnswer format: `x.xxx%`
\end{lstlisting}
\tcblower
\textbf{Answer:} -17.051\%
\end{taskbox}

\begin{taskbox}[title=Law]
\begin{lstlisting}[style=taskstyle]
Using only the Law corpus, compute one normalized dispersion score by joining two groups of materials. For the England-and-Wales reform project, identify the project through its role in recommending changes to the rules governing the two principal ways a body is dealt with after death, including land-capacity pressure affecting one of them; use its original call-for-views document to obtain the opening and closing dates of the response window, and its later summary document to obtain the total submissions received. From the French statutory-code extract pages, retain each distinct code whose territorial-extension/adaptation provisions include a court-label swap from the mainland general court to the locally named first-tier forum. For each retained code, compute the signed calendar-day difference between the footer date tied to the page-copy stamp and the footer date tied to the code-revision stamp. Take the population standard deviation of those signed differences, multiply by the submission total, and divide by both the retained-code count and the inclusive length of the response window. Return only the result rounded to exactly two decimal places, in the form `xx.xx`.
\end{lstlisting}
\tcblower
\textbf{Answer:} 16.87
\end{taskbox}

\begin{taskbox}[title={Technology, Software \& Manufacturing}]
\begin{lstlisting}[style=taskstyle]
A productivity benchmarking team wants one dispersion check for a source-code-line productivity measure in the Japanese development-benchmark family present in this source set. Use the main volume and every companion volume whose preface says it reuses the main volume's analysis layout for business-domain slices. In each selected volume, go to the appendix productivity section where code size is divided by the five-phase development effort. From the row aggregating all size bands and the hourly unit, take the methodology-defined middle statistic rather than the mean, once from the table for first-build projects in the release's bounded analysis window and once from the matching cumulative-period table. Keep the panel only if the number of values equals the length of the contiguous clause range that the toy-safety comparison's counterpart-standard column places outside ordinary-use testing as foreseeable-misuse testing. Apply the benchmark's stated outlier-handling rule, then compute the sample coefficient of variation for the completed panel. Report the result as a percentage rounded to three decimal places. Answer format: `x.xxx%`.
\end{lstlisting}
\tcblower
\textbf{Answer:} 26.747\%
\end{taskbox}

%% file: custom.bib
@article{DBLP:journals/corr/abs-2602-21143,
  author       = {Debjit Paul and
                  Daniel Murphy and
                  Milan Gritta and
                  Ronald Cardenas and
                  Victor Prokhorov and
                  Lena Sophia Bolliger and
                  Aysim Toker and
                  Roy Vaughan Miles and
                  Andreea{-}Maria Oncescu and
                  Jasivan Alex Sivakumar and
                  Philipp Borchert and
                  Ismail Elezi and
                  Meiru Zhang and
                  Ka Yiu Lee and
                  Guchun Zhang and
                  Jun Wang and
                  Gerasimos Lampouras},
  title        = {A Benchmark for Deep Information Synthesis},
  journal      = {CoRR},
  volume       = {abs/2602.21143},
  year         = {2026},
  url          = {https://doi.org/10.48550/arXiv.2602.21143},
  doi          = {10.48550/ARXIV.2602.21143},
  eprinttype   = {arXiv},
  eprint       = {2602.21143},
  bibsource    = {dblp computer science bibliography, https://dblp.org}
}

@article{DBLP:journals/corr/abs-2603-15726,
  author       = {S. Bai and
                  L. Bing and
                  L. Lei and
                  R. Li and
                  X. Li and
                  X. Lin and
                  E. Min and
                  L. Su and
                  B. Wang and
                  L. Wang and
                  L. Wang and
                  S. Wang and
                  X. Wang and
                  Y. Zhang and
                  Z. Zhang and
                  G. Chen and
                  L. Chen and
                  Z. Cheng and
                  Y. Deng and
                  Z. Huang and
                  D. Ng and
                  J. Ni and
                  Q. Ren and
                  X. Tang and
                  B. L. Wang and
                  H. Wang and
                  N. Wang and
                  C. Wei and
                  Q. Wu and
                  J. Xia and
                  Y. Xiao and
                  H. Xu and
                  X. Xu and
                  C. Xue and
                  Z. Yang and
                  Z. Yang and
                  F. Ye and
                  H. Ye and
                  J. Yu and
                  C. Zhang and
                  W. Zhang and
                  H. Zhao and
                  P. Zhu},
  title        = {MiroThinker-1.7 {\&} {H1:} Towards Heavy-Duty Research Agents
                  via Verification},
  journal      = {CoRR},
  volume       = {abs/2603.15726},
  year         = {2026},
  url          = {https://doi.org/10.48550/arXiv.2603.15726},
  doi          = {10.48550/ARXIV.2603.15726},
  eprinttype   = {arXiv},
  eprint       = {2603.15726},
  bibsource    = {dblp computer science bibliography, https://dblp.org}
}

@inproceedings{DBLP:conf/emnlp/ZhengFHCYLL25,
  author       = {Yuxiang Zheng and
                  Dayuan Fu and
                  Xiangkun Hu and
                  Xiaojie Cai and
                  Lyumanshan Ye and
                  Pengrui Lu and
                  Pengfei Liu},
  editor       = {Christos Christodoulopoulos and
                  Tanmoy Chakraborty and
                  Carolyn Rose and
                  Violet Peng},
  title        = {DeepResearcher: Scaling Deep Research via Reinforcement Learning in
                  Real-world Environments},
  booktitle    = {Proceedings of the 2025 Conference on Empirical Methods in Natural
                  Language Processing, {EMNLP} 2025, Suzhou, China, November 4-9, 2025},
  pages        = {414--431},
  publisher    = {Association for Computational Linguistics},
  year         = {2025},
  url          = {https://doi.org/10.18653/v1/2025.emnlp-main.22},
  doi          = {10.18653/V1/2025.EMNLP-MAIN.22},
  bibsource    = {dblp computer science bibliography, https://dblp.org}
}

@article{DBLP:journals/corr/abs-2503-09516,
  author       = {Bowen Jin and
                  Hansi Zeng and
                  Zhenrui Yue and
                  Dong Wang and
                  Hamed Zamani and
                  Jiawei Han},
  title        = {Search-R1: Training LLMs to Reason and Leverage Search Engines with
                  Reinforcement Learning},
  journal      = {CoRR},
  volume       = {abs/2503.09516},
  year         = {2025},
  url          = {https://doi.org/10.48550/arXiv.2503.09516},
  doi          = {10.48550/ARXIV.2503.09516},
  eprinttype   = {arXiv},
  eprint       = {2503.09516},
  bibsource    = {dblp computer science bibliography, https://dblp.org}
}

@inproceedings{DBLP:conf/acl/ZhaoLLZ22,
  author       = {Yilun Zhao and
                  Yunxiang Li and
                  Chenying Li and
                  Rui Zhang},
  editor       = {Smaranda Muresan and
                  Preslav Nakov and
                  Aline Villavicencio},
  title        = {MultiHiertt: Numerical Reasoning over Multi Hierarchical Tabular and
                  Textual Data},
  booktitle    = {Proceedings of the 60th Annual Meeting of the Association for Computational
                  Linguistics (Volume 1: Long Papers), {ACL} 2022, Dublin, Ireland,
                  May 22-27, 2022},
  pages        = {6588--6600},
  publisher    = {Association for Computational Linguistics},
  year         = {2022},
  url          = {https://doi.org/10.18653/v1/2022.acl-long.454},
  doi          = {10.18653/V1/2022.ACL-LONG.454},
  bibsource    = {dblp computer science bibliography, https://dblp.org}
}

@article{DBLP:journals/corr/abs-2601-05163,
  author       = {Qintong Zhang and
                  Xinjie Lv and
                  Jialong Wu and
                  Baixuan Li and
                  Zhengwei Tao and
                  Guochen Yan and
                  Huanyao Zhang and
                  Bin Wang and
                  Jiahao Xu and
                  Haitao Mi and
                  Wentao Zhang},
  title        = {DocDancer: Towards Agentic Document-Grounded Information Seeking},
  journal      = {CoRR},
  volume       = {abs/2601.05163},
  year         = {2026},
  url          = {https://doi.org/10.48550/arXiv.2601.05163},
  doi          = {10.48550/ARXIV.2601.05163},
  eprinttype   = {arXiv},
  eprint       = {2601.05163},
  bibsource    = {dblp computer science bibliography, https://dblp.org}
}

@article{DBLP:journals/corr/abs-2503-13964,
  author       = {Siwei Han and
                  Peng Xia and
                  Ruiyi Zhang and
                  Tong Sun and
                  Yun Li and
                  Hongtu Zhu and
                  Huaxiu Yao},
  title        = {MDocAgent: {A} Multi-Modal Multi-Agent Framework for Document Understanding},
  journal      = {CoRR},
  volume       = {abs/2503.13964},
  year         = {2025},
  url          = {https://doi.org/10.48550/arXiv.2503.13964},
  doi          = {10.48550/ARXIV.2503.13964},
  eprinttype   = {arXiv},
  eprint       = {2503.13964},
  bibsource    = {dblp computer science bibliography, https://dblp.org}
}

@misc{li2025dotsocrmultilingualdocumentlayout,
      title={dots.ocr: Multilingual Document Layout Parsing in a Single Vision-Language Model}, 
      author={Yumeng Li and Guang Yang and Hao Liu and Bowen Wang and Colin Zhang},
      year={2025},
      eprint={2512.02498},
      archivePrefix={arXiv},
      primaryClass={cs.CV},
      url={https://arxiv.org/abs/2512.02498}, 
}

@article{DBLP:journals/corr/abs-2503-21460,
  author       = {Junyu Luo and
                  Weizhi Zhang and
                  Ye Yuan and
                  Yusheng Zhao and
                  Junwei Yang and
                  Yiyang Gu and
                  Bohan Wu and
                  Binqi Chen and
                  Ziyue Qiao and
                  Qingqing Long and
                  Rongcheng Tu and
                  Xiao Luo and
                  Wei Ju and
                  Zhiping Xiao and
                  Yifan Wang and
                  Meng Xiao and
                  Chenwu Liu and
                  Jingyang Yuan and
                  Shichang Zhang and
                  Yiqiao Jin and
                  Fan Zhang and
                  Xian Wu and
                  Hanqing Zhao and
                  Dacheng Tao and
                  Philip S. Yu and
                  Ming Zhang},
  title        = {Large Language Model Agent: {A} Survey on Methodology, Applications
                  and Challenges},
  journal      = {CoRR},
  volume       = {abs/2503.21460},
  year         = {2025},
  url          = {https://doi.org/10.48550/arXiv.2503.21460},
  doi          = {10.48550/ARXIV.2503.21460},
  eprinttype   = {arXiv},
  eprint       = {2503.21460},
  bibsource    = {dblp computer science bibliography, https://dblp.org}
}

@inproceedings{DBLP:conf/emnlp/Yang0ZBCSM18,
  author       = {Zhilin Yang and
                  Peng Qi and
                  Saizheng Zhang and
                  Yoshua Bengio and
                  William W. Cohen and
                  Ruslan Salakhutdinov and
                  Christopher D. Manning},
  editor       = {Ellen Riloff and
                  David Chiang and
                  Julia Hockenmaier and
                  Jun'ichi Tsujii},
  title        = {HotpotQA: {A} Dataset for Diverse, Explainable Multi-hop Question
                  Answering},
  booktitle    = {Proceedings of the 2018 Conference on Empirical Methods in Natural
                  Language Processing, Brussels, Belgium, October 31 - November 4, 2018},
  pages        = {2369--2380},
  publisher    = {Association for Computational Linguistics},
  year         = {2018},
  url          = {https://doi.org/10.18653/v1/d18-1259},
  doi          = {10.18653/V1/D18-1259},
  bibsource    = {dblp computer science bibliography, https://dblp.org}
}

@article{DBLP:journals/tacl/TrivediBKS22,
  author       = {Harsh Trivedi and
                  Niranjan Balasubramanian and
                  Tushar Khot and
                  Ashish Sabharwal},
  title        = {MuSiQue: Multihop Questions via Single-hop Question
                  Composition},
  journal      = {Trans. Assoc. Comput. Linguistics},
  volume       = {10},
  pages        = {539--554},
  year         = {2022},
  url          = {https://doi.org/10.1162/tacl\_a\_00475},
  doi          = {10.1162/TACL\_A\_00475},
  bibsource    = {dblp computer science bibliography, https://dblp.org}
}

@inproceedings{DBLP:conf/naacl/KrishnaKMSSUF25,
  author       = {Satyapriya Krishna and
                  Kalpesh Krishna and
                  Anhad Mohananey and
                  Steven Schwarcz and
                  Adam Stambler and
                  Shyam Upadhyay and
                  Manaal Faruqui},
  editor       = {Luis Chiruzzo and
                  Alan Ritter and
                  Lu Wang},
  title        = {Fact, Fetch, and Reason: {A} Unified Evaluation of Retrieval-Augmented
                  Generation},
  booktitle    = {Proceedings of the 2025 Conference of the Nations of the Americas
                  Chapter of the Association for Computational Linguistics: Human Language
                  Technologies, {NAACL} 2025 - Volume 1: Long Papers, Albuquerque, New
                  Mexico, USA, April 29 - May 4, 2025},
  pages        = {4745--4759},
  publisher    = {Association for Computational Linguistics},
  year         = {2025},
  url          = {https://doi.org/10.18653/v1/2025.naacl-long.243},
  doi          = {10.18653/V1/2025.NAACL-LONG.243},
  bibsource    = {dblp computer science bibliography, https://dblp.org}
}

@inproceedings{DBLP:conf/acl/ZhuLHWZLFC20,
  author       = {Fengbin Zhu and
                  Wenqiang Lei and
                  Youcheng Huang and
                  Chao Wang and
                  Shuo Zhang and
                  Jiancheng Lv and
                  Fuli Feng and
                  Tat{-}Seng Chua},
  editor       = {Chengqing Zong and
                  Fei Xia and
                  Wenjie Li and
                  Roberto Navigli},
  title        = {{TAT-QA:} {A} Question Answering Benchmark on a Hybrid of Tabular
                  and Textual Content in Finance},
  booktitle    = {Proceedings of the 59th Annual Meeting of the Association for Computational
                  Linguistics and the 11th International Joint Conference on Natural
                  Language Processing, {ACL/IJCNLP} 2021, (Volume 1: Long Papers), Virtual
                  Event, August 1-6, 2021},
  pages        = {3277--3287},
  publisher    = {Association for Computational Linguistics},
  year         = {2021},
  url          = {https://doi.org/10.18653/v1/2021.acl-long.254},
  doi          = {10.18653/V1/2021.ACL-LONG.254},
  bibsource    = {dblp computer science bibliography, https://dblp.org}
}

@article{DBLP:journals/corr/abs-2311-11944,
  author       = {Pranab Islam and
                  Anand Kannappan and
                  Douwe Kiela and
                  Rebecca Qian and
                  Nino Scherrer and
                  Bertie Vidgen},
  title        = {FinanceBench: {A} New Benchmark for Financial Question Answering},
  journal      = {CoRR},
  volume       = {abs/2311.11944},
  year         = {2023},
  url          = {https://doi.org/10.48550/arXiv.2311.11944},
  doi          = {10.48550/ARXIV.2311.11944},
  eprinttype   = {arXiv},
  eprint       = {2311.11944},
  bibsource    = {dblp computer science bibliography, https://dblp.org}
}

@inproceedings{DBLP:conf/iclr/MialonF0LS24,
  author       = {Gr{\'{e}}goire Mialon and
                  Cl{\'{e}}mentine Fourrier and
                  Thomas Wolf and
                  Yann LeCun and
                  Thomas Scialom},
  title        = {{GAIA:} a benchmark for General {AI} Assistants},
  booktitle    = {The Twelfth International Conference on Learning Representations,
                  {ICLR} 2024, Vienna, Austria, May 7-11, 2024},
  publisher    = {OpenReview.net},
  year         = {2024},
  url          = {https://openreview.net/forum?id=fibxvahvs3},
  bibsource    = {dblp computer science bibliography, https://dblp.org}
}

@inproceedings{DBLP:conf/iclr/ZhouX0ZLSCOBF0N24,
  author       = {Shuyan Zhou and
                  Frank F. Xu and
                  Hao Zhu and
                  Xuhui Zhou and
                  Robert Lo and
                  Abishek Sridhar and
                  Xianyi Cheng and
                  Tianyue Ou and
                  Yonatan Bisk and
                  Daniel Fried and
                  Uri Alon and
                  Graham Neubig},
  title        = {WebArena: {A} Realistic Web Environment for Building Autonomous Agents},
  booktitle    = {The Twelfth International Conference on Learning Representations,
                  {ICLR} 2024, Vienna, Austria, May 7-11, 2024},
  publisher    = {OpenReview.net},
  year         = {2024},
  url          = {https://openreview.net/forum?id=oKn9c6ytLx},
  bibsource    = {dblp computer science bibliography, https://dblp.org}
}

@article{DBLP:journals/corr/abs-2504-12516,
  author       = {Jason Wei and
                  Zhiqing Sun and
                  Spencer Papay and
                  Scott McKinney and
                  Jeffrey Han and
                  Isa Fulford and
                  Hyung Won Chung and
                  Alex Tachard Passos and
                  William Fedus and
                  Amelia Glaese},
  title        = {BrowseComp: {A} Simple Yet Challenging Benchmark for Browsing Agents},
  journal      = {CoRR},
  volume       = {abs/2504.12516},
  year         = {2025},
  url          = {https://doi.org/10.48550/arXiv.2504.12516},
  doi          = {10.48550/ARXIV.2504.12516},
  eprinttype   = {arXiv},
  eprint       = {2504.12516},
  bibsource    = {dblp computer science bibliography, https://dblp.org}
}

@article{DBLP:journals/corr/abs-2603-08655,
  author       = {Krista Opsahl{-}Ong and
                  Arnav Singhvi and
                  Jasmine Collins and
                  Ivan Zhou and
                  Cindy Wang and
                  Ashutosh Baheti and
                  Owen Oertell and
                  Jacob Portes and
                  Sam Havens and
                  Erich Elsen and
                  Michael Bendersky and
                  Matei Zaharia and
                  Xing Chen},
  title        = {OfficeQA Pro: An Enterprise Benchmark for End-to-End Grounded Reasoning},
  journal      = {CoRR},
  volume       = {abs/2603.08655},
  year         = {2026},
  url          = {https://doi.org/10.48550/arXiv.2603.08655},
  doi          = {10.48550/ARXIV.2603.08655},
  eprinttype   = {arXiv},
  eprint       = {2603.08655},
  bibsource    = {dblp computer science bibliography, https://dblp.org}
}

@inproceedings{DBLP:conf/iclr/YaoZYDSN023,
  author       = {Shunyu Yao and
                  Jeffrey Zhao and
                  Dian Yu and
                  Nan Du and
                  Izhak Shafran and
                  Karthik R. Narasimhan and
                  Yuan Cao},
  title        = {ReAct: Synergizing Reasoning and Acting in Language Models},
  booktitle    = {The Eleventh International Conference on Learning Representations,
                  {ICLR} 2023, Kigali, Rwanda, May 1-5, 2023},
  publisher    = {OpenReview.net},
  year         = {2023},
  url          = {https://openreview.net/forum?id=WE\_vluYUL-X},
  bibsource    = {dblp computer science bibliography, https://dblp.org}
}

@inproceedings{DBLP:conf/nips/SchickDDRLHZCS23,
  author       = {Timo Schick and
                  Jane Dwivedi{-}Yu and
                  Roberto Dess{\`{\i}} and
                  Roberta Raileanu and
                  Maria Lomeli and
                  Eric Hambro and
                  Luke Zettlemoyer and
                  Nicola Cancedda and
                  Thomas Scialom},
  editor       = {Alice Oh and
                  Tristan Naumann and
                  Amir Globerson and
                  Kate Saenko and
                  Moritz Hardt and
                  Sergey Levine},
  title        = {Toolformer: Language Models Can Teach Themselves to Use Tools},
  booktitle    = {Advances in Neural Information Processing Systems 36: Annual Conference
                  on Neural Information Processing Systems 2023, NeurIPS 2023, New Orleans,
                  LA, USA, December 10 - 16, 2023},
  year         = {2023},
  url          = {http://papers.nips.cc/paper\_files/paper/2023/hash/d842425e4bf79ba039352da0f658a906-Abstract-Conference.html},
  bibsource    = {dblp computer science bibliography, https://dblp.org}
}

@inproceedings{DBLP:conf/nips/YangJWLYNP24,
  author       = {John Yang and
                  Carlos E. Jimenez and
                  Alexander Wettig and
                  Kilian Lieret and
                  Shunyu Yao and
                  Karthik Narasimhan and
                  Ofir Press},
  editor       = {Amir Globersons and
                  Lester Mackey and
                  Danielle Belgrave and
                  Angela Fan and
                  Ulrich Paquet and
                  Jakub M. Tomczak and
                  Cheng Zhang},
  title        = {SWE-agent: Agent-Computer Interfaces Enable Automated Software Engineering},
  booktitle    = {Advances in Neural Information Processing Systems 38: Annual Conference
                  on Neural Information Processing Systems 2024, NeurIPS 2024, Vancouver,
                  BC, Canada, December 10 - 15, 2024},
  year         = {2024},
  url          = {http://papers.nips.cc/paper\_files/paper/2024/hash/5a7c947568c1b1328ccc5230172e1e7c-Abstract-Conference.html},
  bibsource    = {dblp computer science bibliography, https://dblp.org}
}

@inproceedings{DBLP:conf/iclr/0001LSXTZPSLSTL25,
  author       = {Xingyao Wang and
                  Boxuan Li and
                  Yufan Song and
                  Frank F. Xu and
                  Xiangru Tang and
                  Mingchen Zhuge and
                  Jiayi Pan and
                  Yueqi Song and
                  Bowen Li and
                  Jaskirat Singh and
                  Hoang H. Tran and
                  Fuqiang Li and
                  Ren Ma and
                  Mingzhang Zheng and
                  Bill Qian and
                  Yanjun Shao and
                  Niklas Muennighoff and
                  Yizhe Zhang and
                  Binyuan Hui and
                  Junyang Lin and
                  et al.},
  title        = {OpenHands: An Open Platform for {AI} Software Developers as Generalist
                  Agents},
  booktitle    = {The Thirteenth International Conference on Learning Representations,
                  {ICLR} 2025, Singapore, April 24-28, 2025},
  publisher    = {OpenReview.net},
  year         = {2025},
  url          = {https://openreview.net/forum?id=OJd3ayDDoF},
  bibsource    = {dblp computer science bibliography, https://dblp.org}
}

@inproceedings{DBLP:conf/nips/ZhengYXS0YCKSGB24,
  author       = {Lianmin Zheng and
                  Liangsheng Yin and
                  Zhiqiang Xie and
                  Chuyue Sun and
                  Jeff Huang and
                  Cody Hao Yu and
                  Shiyi Cao and
                  Christos Kozyrakis and
                  Ion Stoica and
                  Joseph E. Gonzalez and
                  Clark W. Barrett and
                  Ying Sheng},
  editor       = {Amir Globersons and
                  Lester Mackey and
                  Danielle Belgrave and
                  Angela Fan and
                  Ulrich Paquet and
                  Jakub M. Tomczak and
                  Cheng Zhang},
  title        = {SGLang: Efficient Execution of Structured Language Model Programs},
  booktitle    = {Advances in Neural Information Processing Systems 38: Annual Conference
                  on Neural Information Processing Systems 2024, NeurIPS 2024, Vancouver,
                  BC, Canada, December 10 - 15, 2024},
  year         = {2024},
  url          = {http://papers.nips.cc/paper\_files/paper/2024/hash/724be4472168f31ba1c9ac630f15dec8-Abstract-Conference.html},
  bibsource    = {dblp computer science bibliography, https://dblp.org}
}

@article{DBLP:journals/corr/abs-2505-15117,
  author       = {Bowen Jin and
                  Jinsung Yoon and
                  Priyanka Kargupta and
                  Sercan {\"{O}}. Arik and
                  Jiawei Han},
  title        = {An Empirical Study on Reinforcement Learning for Reasoning-Search
                  Interleaved {LLM} Agents},
  journal      = {CoRR},
  volume       = {abs/2505.15117},
  year         = {2025},
  url          = {https://doi.org/10.48550/arXiv.2505.15117},
  doi          = {10.48550/ARXIV.2505.15117},
  eprinttype   = {arXiv},
  eprint       = {2505.15117},
  bibsource    = {dblp computer science bibliography, https://dblp.org}
}

@inproceedings{DBLP:conf/nips/XieZCLZCHCSLLXZ24,
  author       = {Tianbao Xie and
                  Danyang Zhang and
                  Jixuan Chen and
                  Xiaochuan Li and
                  Siheng Zhao and
                  Ruisheng Cao and
                  Toh Jing Hua and
                  Zhoujun Cheng and
                  Dongchan Shin and
                  Fangyu Lei and
                  Yitao Liu and
                  Yiheng Xu and
                  Shuyan Zhou and
                  Silvio Savarese and
                  Caiming Xiong and
                  Victor Zhong and
                  Tao Yu},
  editor       = {Amir Globersons and
                  Lester Mackey and
                  Danielle Belgrave and
                  Angela Fan and
                  Ulrich Paquet and
                  Jakub M. Tomczak and
                  Cheng Zhang},
  title        = {OSWorld: Benchmarking Multimodal Agents for Open-Ended Tasks in Real
                  Computer Environments},
  booktitle    = {Advances in Neural Information Processing Systems 38: Annual Conference
                  on Neural Information Processing Systems 2024, NeurIPS 2024, Vancouver,
                  BC, Canada, December 10 - 15, 2024},
  year         = {2024},
  url          = {http://papers.nips.cc/paper\_files/paper/2024/hash/5d413e48f84dc61244b6be550f1cd8f5-Abstract-Datasets\_and\_Benchmarks\_Track.html},
  bibsource    = {dblp computer science bibliography, https://dblp.org}
}

@inproceedings{DBLP:conf/nips/MaZZYZZLW024,
  author       = {Zeyao Ma and
                  Bohan Zhang and
                  Jing Zhang and
                  Jifan Yu and
                  Xiaokang Zhang and
                  Xiaohan Zhang and
                  Sijia Luo and
                  Xi Wang and
                  Jie Tang},
  editor       = {Amir Globersons and
                  Lester Mackey and
                  Danielle Belgrave and
                  Angela Fan and
                  Ulrich Paquet and
                  Jakub M. Tomczak and
                  Cheng Zhang},
  title        = {SpreadsheetBench: Towards Challenging Real World Spreadsheet Manipulation},
  booktitle    = {Advances in Neural Information Processing Systems 38: Annual Conference
                  on Neural Information Processing Systems 2024, NeurIPS 2024, Vancouver,
                  BC, Canada, December 10 - 15, 2024},
  year         = {2024},
  url          = {http://papers.nips.cc/paper\_files/paper/2024/hash/ac840df270ac537dd74530a15c332684-Abstract-Datasets\_and\_Benchmarks\_Track.html},
  bibsource    = {dblp computer science bibliography, https://dblp.org}
}

@inproceedings{DBLP:conf/emnlp/LinLZZLWT25,
  author       = {Teng Lin and
                  Yuyu Luo and
                  Honglin Zhang and
                  Jicheng Zhang and
                  Chunlin Liu and
                  Kaishun Wu and
                  Nan Tang},
  editor       = {Christos Christodoulopoulos and
                  Tanmoy Chakraborty and
                  Carolyn Rose and
                  Violet Peng},
  title        = {MEBench: Benchmarking Large Language Models for Cross-Document Multi-Entity
                  Question Answering},
  booktitle    = {Proceedings of the 2025 Conference on Empirical Methods in Natural
                  Language Processing, {EMNLP} 2025, Suzhou, China, November 4-9, 2025},
  pages        = {1481--1494},
  publisher    = {Association for Computational Linguistics},
  year         = {2025},
  url          = {https://doi.org/10.18653/v1/2025.emnlp-main.77},
  doi          = {10.18653/V1/2025.EMNLP-MAIN.77},
  bibsource    = {dblp computer science bibliography, https://dblp.org}
}

@article{DBLP:journals/corr/abs-2407-19056,
  author       = {Zilong Wang and
                  Yuedong Cui and
                  Li Zhong and
                  Zimin Zhang and
                  Da Yin and
                  Bill Yuchen Lin and
                  Jingbo Shang},
  title        = {OfficeBench: Benchmarking Language Agents across Multiple Applications
                  for Office Automation},
  journal      = {CoRR},
  volume       = {abs/2407.19056},
  year         = {2024},
  url          = {https://doi.org/10.48550/arXiv.2407.19056},
  doi          = {10.48550/ARXIV.2407.19056},
  eprinttype   = {arXiv},
  eprint       = {2407.19056},
  bibsource    = {dblp computer science bibliography, https://dblp.org}
}

@inproceedings{DBLP:conf/coling/HoNSA20,
  author       = {Xanh Ho and
                  Anh{-}Khoa Duong Nguyen and
                  Saku Sugawara and
                  Akiko Aizawa},
  editor       = {Donia Scott and
                  N{\'{u}}ria Bel and
                  Chengqing Zong},
  title        = {Constructing {A} Multi-hop {QA} Dataset for Comprehensive Evaluation
                  of Reasoning Steps},
  booktitle    = {Proceedings of the 28th International Conference on Computational
                  Linguistics, {COLING} 2020, Barcelona, Spain (Online), December 8-13,
                  2020},
  pages        = {6609--6625},
  publisher    = {International Committee on Computational Linguistics},
  year         = {2020},
  url          = {https://doi.org/10.18653/v1/2020.coling-main.580},
  doi          = {10.18653/V1/2020.COLING-MAIN.580},
  bibsource    = {dblp computer science bibliography, https://dblp.org}
}

@article{DBLP:journals/corr/abs-2401-15391,
  author       = {Yixuan Tang and
                  Yi Yang},
  title        = {MultiHop-RAG: Benchmarking Retrieval-Augmented Generation for Multi-Hop
                  Queries},
  journal      = {CoRR},
  volume       = {abs/2401.15391},
  year         = {2024},
  url          = {https://doi.org/10.48550/arXiv.2401.15391},
  doi          = {10.48550/ARXIV.2401.15391},
  eprinttype   = {arXiv},
  eprint       = {2401.15391},
  bibsource    = {dblp computer science bibliography, https://dblp.org}
}

@inproceedings{DBLP:conf/emnlp/LiS0L0C24,
  author       = {Chuhan Li and
                  Ziyao Shangguan and
                  Yilun Zhao and
                  Deyuan Li and
                  Yixin Liu and
                  Arman Cohan},
  editor       = {Yaser Al{-}Onaizan and
                  Mohit Bansal and
                  Yun{-}Nung Chen},
  title        = {M3SciQA: {A} Multi-Modal Multi-Document Scientific {QA} Benchmark
                  for Evaluating Foundation Models},
  booktitle    = {Findings of the Association for Computational Linguistics: {EMNLP}
                  2024, Miami, Florida, USA, November 12-16, 2024},
  series       = {Findings of {ACL}},
  pages        = {15419--15446},
  publisher    = {Association for Computational Linguistics},
  year         = {2024},
  url          = {https://doi.org/10.18653/v1/2024.findings-emnlp.904},
  doi          = {10.18653/V1/2024.FINDINGS-EMNLP.904},
  bibsource    = {dblp computer science bibliography, https://dblp.org}
}

@article{DBLP:journals/corr/abs-2510-04374,
  author       = {Tejal Patwardhan and
                  Rachel Dias and
                  Elizabeth Proehl and
                  Grace Kim and
                  Michele Wang and
                  Olivia Watkins and
                  Sim{\'{o}}n Posada Fishman and
                  Marwan Aljubeh and
                  Phoebe Thacker and
                  Laurance Fauconnet and
                  Natalie S. Kim and
                  Patrick Chao and
                  Samuel Miserendino and
                  Gildas Chabot and
                  David Li and
                  Michael Sharman and
                  Alexandra Barr and
                  Amelia Glaese and
                  Jerry Tworek},
  title        = {GDPval: Evaluating {AI} Model Performance on Real-World Economically
                  Valuable Tasks},
  journal      = {CoRR},
  volume       = {abs/2510.04374},
  year         = {2025},
  url          = {https://doi.org/10.48550/arXiv.2510.04374},
  doi          = {10.48550/ARXIV.2510.04374},
  eprinttype   = {arXiv},
  eprint       = {2510.04374},
  bibsource    = {dblp computer science bibliography, https://dblp.org}
}

@misc{openai2026gpt55systemcard,
  author      = {OpenAI},
  title       = {{GPT}-5.5 {S}ystem {C}ard},
  year        = {2026},
  url         = {https://deploymentsafety.openai.com/gpt-5-5},
}

@misc{google2026gemini31prosystemcard,
  author      = {Google},
  title       = {{Gemini}-3.1 {P}ro {S}ystem {C}ard},
  year        = {2026},
  url         = {https://deepmind.google/models/model-cards/gemini-3-1-pro/},
}

@manual{deepseek2026deepseekv4techreport,
  author      = {DeepSeek-AI},
  title       = {{DeepSeek}-{V}4 {T}echnical {R}eport},
  year        = {2026},
  url         = {https://huggingface.co/deepseek-ai/DeepSeek-V4-Pro/blob/main/DeepSeek_V4.pdf},
}

@misc{qwen35blog,
    title = {Qwen3.5: Accelerating Productivity with Native Multimodal Agents},
    url = {https://qwen.ai/blog?id=qwen3.5},
    author = {{Qwen Team}},
    month = {February},
    year = {2026}
}

@misc{google2026gemini31flashlitesystemcard,
  author      = {Google},
  title       = {{Gemini}-3.1 {F}lash-{L}ite {S}ystem {C}ard},
  year        = {2026},
  url         = {https://deepmind.google/models/model-cards/gemini-3-1-flash-lite/},
}

@misc{zhipu2026glm51systemcard,
  author      = {Zhipu},
  title       = {{GLM}-5.1 {S}ystem {C}ard},
  year        = {2026},
  url         = {https://docs.bigmodel.cn/cn/guide/models/text/glm-5.1},
}
